*Article*

# Indoor Localization for Personalized Ambient Assisted Living of Multiple Users in Multi-Floor Smart Environments

Nirmalya Thakur * and Chia Y. Han

Department of Electrical Engineering and Computer Science, University of Cincinnati, Cincinnati, OH 45221-0030, USA; han@ucmail.uc.edu
* Correspondence: thakurna@mail.uc.edu

**Abstract:** This paper presents a multifunctional interdisciplinary framework that makes four scientific contributions towards the development of personalized ambient assisted living (AAL), with a specific focus to address the different and dynamic needs of the diverse aging population in the future of smart living environments. First, it presents a probabilistic reasoning-based mathematical approach to model all possible forms of user interactions for any activity arising from user diversity of multiple users in such environments. Second, it presents a system that uses this approach with a machine learning method to model individual user-profiles and user-specific user interactions for detecting the dynamic indoor location of each specific user. Third, to address the need to develop highly accurate indoor localization systems for increased trust, reliance, and seamless user acceptance, the framework introduces a novel methodology where two boosting approaches—Gradient Boosting and the AdaBoost algorithm are integrated and used on a decision tree-based learning model to perform indoor localization. Fourth, the framework introduces two novel functionalities to provide semantic context to indoor localization in terms of detecting each user's floor-specific location as well as tracking whether a specific user was located inside or outside a given spatial region in a multi-floor-based indoor setting. These novel functionalities of the proposed framework were tested on a dataset of localization-related Big Data collected from 18 different users who navigated in 3 buildings consisting of 5 floors and 254 indoor spatial regions, with an to address the limitation in prior works in this field centered around the lack of training data from diverse users. The results show that this approach of indoor localization for personalized AAL that models each specific user always achieves higher accuracy as compared to the traditional approach of modeling an average user. The results further demonstrate that the proposed framework outperforms all prior works in this field in terms of functionalities, performance characteristics, and operational features.

## 1. Introduction

Today, the longevity of people has increased worldwide, and most individuals may expect to survive until they are 60 or older for the first time in history on account of advances in healthcare and medical research [1]. The aging population across the world, currently at 962 million [2], is increasing at an unprecedented rate and is projected to reach 2 billion by 2050 [3]. The aging population is associated with diverse and dynamic needs on account of different rates of decline in behavioral, social, emotional, mental, psychological, and motor abilities, as well as other issues such as cognitive impairment, behavioral disorders, disabilities, neurological disorders, Dementia, Alzheimer's, and visual impairments, that are commonly associated with aging and have different effects on the elderly based on their diversity [4]. This increase in the aging population of the world has been accompanied by a decrease in the caregiver population to look after the elderly [5,6], which has created a multitude of problems [7–9]. We outline two of the major problems here. First, due to the increase in demand, the cost of caregiving has significantly increased in the last few years. Therefore, affording caregivers is becoming a concern for middle-income and low-income families. Second, caregivers are quite often fatigued, overworked, overwhelmed, and overburdened as they have to look after the needs of multiple older adults with different needs arising from various kinds of limitations associated with old age. This leads to a decrease in the quality of care.

As living infrastructure changes and society advances, we anticipate that a majority of the world's population, including the elderly, will reside in interconnected Smart Homes, Smart Communities, and Smart Cities in the near future [10]. In fact, a recent study [11] has projected that by the year 2050, approximately 68% of the global population will be living in Internet of Things (IoT)-based Smart Homes. Thus, in view of the shortage of caregivers and this projected increase in Smart Home adoption, it is the need of the hour that the future of technology-based smart living environments such as smart homes can contribute towards Ambient Assisted Living (AAL) of the elderly by being able to detect, interpret, analyze, and anticipate their varying needs in the context of their user interactions with various forms of technology-based systems in such living environments [12]. AAL may be broadly defined as the use of interconnected technology-based automated and semi-automated applications and solutions in a person's living environment that can communicate and interact with each other with an aim to improve their health and well-being, quality of life, and independence in the context of their interactions with such environments [13].

Despite recent research and development in the field of AAL, as discussed in detail in Section 2, several challenges still exist in the context of the development of such AAL functionalities in Smart Homes. These are outlined as follows:

1. Most of the advancement in AAL research has focused on the development of technologies based on what is feasible [14] and by keeping an average user or in other words an 'one size fits all' approach in mind [15,16]. Researchers have defined an average user in terms of specific patterns of user interactions and specific cognitive, behavioral, perceptual, emotional, and mental abilities, which are quite often different from the actual user, who might present different needs and varying abilities based on their diversity. Recent research [17] shows that such approaches are no longer effective as specific users present specific needs [18,19].
2. The attempts [16] to customize such applications for specific needs of the actual user have focused on manual redesign of the systems based on the individual needs, training the actual user to interact like the average user based on whom the system was initially designed, and supplying the user with a different or additional gadget or tool to help them with the interactions. Such customization initiatives are complicated to develop, expensive, involve a significant amount of time to implement, and are not practically feasible in most cases. Moreover, the elderly, being naturally receptive to new technologies [20], quite often refuse to use a different or additional gadget or tool for their daily interactions.
3. The differences in the anticipated user interactions by these AAL systems based on the model of an average user and the actual interactions by the users based on their specific needs creates a 'gap' in terms of the effectiveness of these systems to respond to these varying needs. This creates perceptions of complexity, fear, anxiety, lack of trust in technology, and confusion in the mind of the actual user, who ultimately refuses to use the given AAL system or tool [16,21,22].
4. The AAL technologies developed by keeping an average user in mind have not considered the dynamic changes in specific needs of actual users that could be demonstrated on a temporary basis, such as from an injury [16].
5. Smart Homes of the future would involve multiple users, including the elderly, interacting, and living with each other [23,24]. These users are expected to be diverse in multiple ways [25]. User diversity presents a challenge in terms of multi-user activity analysis or for analyzing any associated needs of the users in their living environments [26].

In view of the above challenges, it may be concluded that the future of AAL technologies in Smart Homes should be developed to have a "personalized" touch so that these technologies are able to respond, address, anticipate, and adapt to the diverse and dynamic needs, requirements, and challenges faced by actual users at specific points of time by tracking the underlining user interactions.

The different kinds of needs in the elderly in their living environments that creates the necessity for developing AAL systems with a "personalized" touch are mainly associated with the requirements and challenges they face in completing activities of daily living. Being able to perform activities of daily living in an independent manner is crucial for sustenance, improved quality of life, and healthy aging for the elderly [27]. Activities of daily living are characterized by a diverse set of user interactions based on the diversity of the user as well as the variations in the environment parameters or context attributes in the confines of the indoor spatial environment of the user. User interactions with one or more environment variables or context parameters may be broadly summarized as the user responding to a need, motive, objective, or goal [28], which is a function of the user's behavioral, navigational, and localization-related characteristics, that further depend on the multiple characteristics of the user and the features of the environment [29,30]. Tracking and studying these behavioral, navigational, and localization-related characteristics, distinct to each user in a given environment, through user-specific indoor localization, therefore, holds the potential towards addressing the challenges for the creation of personalized AAL living experiences for multiple users in the future of smart living spaces, such as Smart Homes [31]. In a broad manner, an Indoor Localization System may be defined as a host of technologies, networks, and devices that communicate with each other to track and locate individuals, objects, and assets in indoor environments. Here, by indoor environments, we refer to fully or partially closed indoor settings where navigation technologies such as Global Positioning Systems (GPS) and Global Navigation Satellite Systems (GNSS) fail to work [32]. This demonstrates the potential of indoor localization towards the creation of "personalized" AAL experiences in the future of Smart Homes for the healthy aging and independent living of the diverse aging population. Therefore, this paper proposes a multifunctional framework for personalized AAL for multiple users through the lens of indoor localization. This framework was developed by integrating the latest advancements from Pervasive Computing, Big Data, Information Retrieval, Internet of Things, Human–Computer Interaction, Machine Learning, and Pattern Recognition.

This paper is organized as follows. We present an overview of related works in this field in Section 2. The methodology and system architecture of this framework is presented in Section 3. Section 4 discusses the results and findings, which is followed by a comparative discussion in Section 5 that upholds the relevance and significance of the framework and discusses how it outperforms similar works in this field. The conclusion and scope for future work are presented in Section 6.

## 2. Literature Review

In this section, we present a comprehensive review of recent works in the field of Indoor Localization that focus on Ambient Assisted Living. These works involved development of a range of approaches and methodologies at the intersection of different disciplines to detect the indoor location of the user. Some of these approaches were evaluated by use of datasets, and the rest were evaluated by data collected from real-time or simulated experiments.

Varma et al. [33] used a Random Forest-driven machine learning classifier to develop an indoor localization system that used the data from 13 beacons that were set up in a simulated Internet of Things (IoT)-based environment. The data from these beacons were interpreted to obtain the user's position. Gao et al. [34] used signals coming from a Wi-Fi to train a Random Forest-based learning approach. The system used the concept of region-based division of location grids to reduce the induced error and to detect the user in a specific grid. Khan et al. [35] trained an artificial neural network by using the data coming from Wireless Local Area Network (WLAN) access points and Wireless Sensor Networks (WSNs) to detect the indoor position of the user. Labinghisa et al. [36] used the methodology of virtual access points-based mapping for training an artificial neural network for indoor location detection. This methodology had the advantage that more access points could be used without installing any new hardware. A neural network-based Wi-Fi fingerprinting approach was developed by Qin et al. [37] to detect the location of a user in an indoor environment. The authors evaluated their work on two datasets to uphold the relevance of the same. A decision tree-based approach was used by Musa et al. [38] to detect the indoor location of a user. The system architecture consisted of using a non-line of sight approach and a multipath propagation tracking while using the ultra-wide band methodology. A similar approach that was also decision tree-driven was proposed by Yim et al. [39]. This system had the functionality to train the decision tree in the off-line phase, which used the data from Wi-Fi fingerprinting. In the approach proposed by Sjoberg et al. [40], visual features of a given environment were used to develop a bag-of-words, which was used to train a support vector machine (SVM) learner. Zhang et al. [41] proposed a 2.5D indoor positioning system that used the data from Wi-Fi signals and the user's altitude to train an SVM classifier.

A k-NN classifier-based approach that used the signal strength fingerprint technology for indoor localization was proposed by Zhang et al. [42]. Ge et al. [43] developed an algorithm to detect and interpret the data obtained from access points by using a k-NN-based learning method to detect the indoor position of the user. In the methodology proposed by Hu et al. [44], the k-NN classifier detected the indoor location of the user based on the nearest access point to the user. The primary finding of Hu et al.'s work was that the highest performance accuracy was observed for k = 1. A 3D positioning methodology was proposed by Zhang et al. [45] for a hospital setting. It used the cellular network data as well and the data from Wi-Fi access points to detect the latitude and longitude of the user's indoor location by using a deep learning approach. Another deep learning-based approach for indoor location detection was proposed by Poulose et al. [46] that used the data from RSSI signals to train the learning model. Jamâa et al. [47] proposed an approach that involved training a linear regression-based learning model that performed distance-based analysis to detect the indoor location of a user. The system used a methodology where each anchor node in the given environment had its own linear ranging approach. Barsocchi et al. [48] used a linear regression-based learning approach to develop an indoor positioning system that tracked the user's distance from different reference points by analyzing the RSSI values, and thereafter it mapped the numerical value of the same to a distance measure to detect the actual position of the user.

Researchers in this field have focused on developing indoor localization systems by tracking the X and Y coordinates of the user's indoor location. In [49], Bolic et al. developed a new RFID-based indoor location tracking device that could detect proximity tags in the user's environment to detect the user's location with a Root Mean Squared Error (RMSE) of 0.32 m. A Bayesian approach was used by Angermann et al. [50] for tracking the location of pedestrians in terms of their spatial coordinates in indoor settings with an RMSE of 1–2 m. Evennou et al. [51] developed a framework that used signal processing concepts to interpret user interaction data for detecting the coordinate-based location of the user with an RMSE of 1.53 m. The methodology proposed by Wang et al. [52] involved using particle filters and WLAN approaches to track the location information of the user in terms of spatial coordinates. The system had an RMSE of 4.3 m. Klingbeil et al. [53] proposed an approach for detecting the X and Y coordinates of a user's indoor location that was Monte Carlo-based and had an RMSE of 1.2 m.

Several works in the field of indoor localization have involved testing and evaluation of the proposed approaches in real-time with multiple participants who were either recruited or volunteered to participate in the experimental trials. Murata et al. [54] proposed a smartphone-driven system that used the data from RSSI sensors and BLE beacons to detect the indoor location of the user, which was evaluated by collecting the data from 10 participants. Yoo et al. [55] used a Gaussian approach to develop an indoor localization framework that used the concept of trajectory learning with crowdsourced measurements to implement localization in the absence of a map. The authors tested their approach by recruiting 10 participants. Kothari et al. [56] developed a smartphone-driven approach that was cost-effective and was able to detect the indoor location of a user with high accuracy. The approach combined concepts from dead reckoning and Wi-Fi fingerprinting and was tested by including 4 participants in the experimental trials. In the work proposed by Qian et al. [57], concepts of stride length estimation and step detection were implemented by using a principal component approach (PCA) and a pedestrian reckoning (PDR) approach. The data collected from a total of 3 participants were used to discuss the performance characteristics of the system. Fusco et al. [58] proposed a framework that used the data collected from a smartphone which was interpreted and analyzed by methodologies from computer vision and visual-inertial odometry to detect the indoor location of a user. The data collection and performance evaluation process included 3 participants. Chang et al. [59] recruited 3 participants to validate their indoor localization framework that was deep neural network-driven and was trained using Wi-Fi channel state information data obtained from one access point in the premises of the user to detect the user's indoor location. Subbu et al. [60] developed an indoor positioning system that used data collected from a nexus one smartphone. The system collected the magnetic field data from this smartphone and classified the magnetic signatures using dynamic time warping to detect the user's position. A total of 4 participants participated in the experimental trials. Zhou et al. [61] developed an activity recognition-based indoor location detection system that could detect 9 different activities and the associated locations in the indoor environment of the user where these respective activities were performed. The authors tested their approach by including 10 participants in the experiments.

Chen et al. [62] proposed a Multi-dimensional Dynamic Time Warping (MDTW) approach to analyze the Wi-Fi fingerprint data by using the least-squares methodology. The system was evaluated by collecting and analyzing user interaction data from 6 participants. Xu et al. [63] proposed ClickLoc, which was a computer vision-based indoor localization system that used bootstrapping concepts and a place of interest (POI) approach to detect the location of a user. Data collected from 2 participants were used to evaluate the performance characteristics of the system. In the indoor positioning approach proposed by Wang et al. [64], dead reckoning approaches were used to track the position of a user in terms of environment signatures and internal landmarks that were specific to a given environment. The authors discussed the performance characteristics of their approach by taking into consideration the data collected from 3 participants. Röbesaat et al. [65] used Kalman filtering-based fusion to develop a system that tracked the data collected from an android device and BLE modules to detect the indoor location of a user, which was tested by including 4 participants in the study. Yang et al. [66] developed an acoustic communication-based framework that allowed users to share and exchange their locations in an indoor environment when they met with each other. The authors conducted experiments by including 10 participants in the trials to validate their approach. Wu et al. [67] developed an indoor localization system that used the data from novel sensors that were integrated into mobile phones of users that could leverage characteristics of user motion and user behavior to develop a radio map of a floor plan—which could then be used to detect the indoor locations of the users. A total of 4 participants participated in the study. Gu et al. [68] proposed a step counting algorithm that could address challenges such as overcounting of steps and false walking while tracking the indoor location of the user; that was validated by taking into consideration the data collected from 8 participants. In [69], Niu et al. proposed OWUH, an Online Learning-based WIFI Radio Map Updating service that worked by combining old and new RSSI data and probe data to increment and update the radio map of the given environment. This functionality allowed the system to work with a smaller number of RSSI datapoints while still being able to detect the indoor location of the user with a high accuracy that was deduced by the authors from the results of their experimental trials that included 15 participants. In addition to the above, recent works in this field have focused on building tailored magnetic maps for smartphones [70,71] and using edge computing approaches to analyze environmental sensor data [72,73] for indoor localization. While a few recent works [74–80] have investigated approaches for floor detection in the context of indoor localization but the performance characteristics of such systems are not that high to support widescale deployment and real-time implementation of the same.

Despite these recent works in the field of AAL, there exist several limitations and challenges that are yet to be addressed in this field of research. These include:

1. The works in AAL [54–69] that involved recruiting participants to evaluate the efficacy of the proposed approaches were developed by keeping an average user in mind [16], where the average user was defined to have a certain set of user interaction patterns in terms of specific cognitive, behavioral, perceptual, and mental abilities, which in a real-world scenario can be different from the characteristics, needs, and abilities of the actual user in the AAL environment.
2. The works [33–48] that used different forms of machine learning and artificial intelligence approaches to detect the indoor location of a user have used the major machine learning approaches and did not focus on using any form of boosting approaches to improve the performance accuracy of the underlining systems. To improve the trust and seamless acceptance of such AAL technologies as well as to contribute towards improved quality of life and enhanced user experience of the elderly, it is crucial to improve the performance accuracies of such systems.
3. The approaches [54–69] for indoor localization, which were evaluated by including multiple participants or users in the experimental trials, did not have a significant number of participants to represent the diversity of actual users [25]. It is important to include more participants in such experimental trials so that the machine learning-based systems can get familiar with the diversified range of user interactions from different users in the real world.
4. The indoor localization frameworks [49–53] that focused on detecting the X and Y coordinates of the user's indoor position did not focus on providing semantic context to these detections. Here, semantic context refers to providing additional meaning and details in terms of building, floor, and dynamic spatial context information (such as inside or outside a given indoor spatial region) to such indoor location detections, for better understanding and interpretation of the indoor locations of the user in real world scenarios; where the users could be living in a multi-storied building, so that the same may be interpreted and analyzed for immediate care and attention in case of any healthcare-related needs. While a few recent works [74–80] have investigated approaches for floor detection in indoor localization, the performance accuracies of such systems are not high enough to support their widescale deployment and real time implementations.

Addressing these challenges by integrating the latest advancements from Pervasive Computing, Big Data, Information Retrieval, Internet of Things, Human–Computer Interaction, Machine Learning, and Pattern Recognition with a specific focus on developing a "personalized" touch in the future of multi-user-based AAL environments through the lens of Indoor Localization serves as the main motivation for the development of this framework, which is introduced and discussed in detail in Section 3.

## 3. Proposed Work

This section presents the steps towards the development of the proposed framework. In a real-world scenario, human activities are complex and characterized by diverse behavioral, navigational, and movement patterns that depend on both external and internal factors related to the user performing these activities. Here, external factors broadly refer to the characteristic features and context attributes of the environment in which these specific activities are performed, and internal factors refer to the abilities of the user in terms of physical, mental, cognitive, psychological, and emotional, just to name a few that determine the user's performance during such activities [28–30]. Such activities that are performed in the real world are known as complex activities and comprise of atomic activities, context attributes, core atomic activities, core context attributes, other atomic activities, other context attributes, start atomic activities, end atomic activities, start context attributes, and end context attributes [81], which can be different for different users based on these internal and external factors affecting the activity being performed. As per [81], atomic activities are defined as the granular-level tasks and actions associated with the complex activity, and the parameters of the environment that are associated with the user interactions related to these atomic activities are known as context attributes. Those granular-level tasks and actions that represent the beginning of an activity and end of an activity are known as the start atomic activities and end atomic activities, respectively. The environment parameters that are involved in the user interactions of the start and end atomic activities are known as start context attributes and end context attributes, respectively. Those atomic activities that are crucial for the completion of a given activity are known as core atomic activities, and the context attributes on which these core atomic activities take place are known as core context attributes.

As different users, based on their diversity and the associated internal and external factors, can exhibit different forms of navigation and movement patterns during performing different activities in a given environment, that are crucial for personalized indoor localization, it is important to model the diverse ways in which these activities can be performed by different users. To achieve the same, we use the probabilistic reasoning-based mathematical model proposed in [82] that presents multiple equations to model these different ways by which a complex activity may be performed. These equations, as shown in Equations (1)–(3), are based on the concept of analyzing a complex activity in terms of atomic activities, context attributes, other atomic activities, other context attributes, start atomic activities, end atomic activities, start context attributes, and end context attributes.

$$\zeta(t) = A_t C0 + A_t C1 + A_t C2 + \ldots\ldots A_t CA_t = 2^{A_t} \qquad (1)$$

$$\Theta(t) = (A_t - c_t)C0 + (A_t - c_t)C1 + (A_t - c_t)C2 + .. + (A_t - c_t)C(A_t - c_t) = 2^{(A_t - c_t)} \qquad (2)$$

$$\Psi(t) = 2^{A_t} - 2^{(A_t - c_t)} = 2^{(A_t - c_t)} * (2^{c_t} - 1) \qquad (3)$$

where:

$\zeta(t)$: all the different ways in which a complex activity can be performed by different users

$\Theta(t)$: all the different ways in which a complex activity can be performed where the specific user performing the activity always reaches the end goal

$\Psi(t)$: all the different ways of performing a complex activity where the specific user performing the activity never reaches the end goal

$A_t$: atomic activities for a complex activity

$C_t$: context attributes for a complex activity

$A_{ts}$: all the start atomic activities for a given complex activity

$C_{ts}$: all the start context attributes for a given complex activity

$A_{tE}$: all the end atomic activities for a given complex activity

$C_{tE}$: all the end context attributes for a given complex activity

$A_{t\delta}$: all the core atomic activities for a given complex activity

$C_{t\delta}$: all the core context attributes for a given complex activity

$A_{tI}$: all the atomic activities for a given complex activity
$C_{tI}$: all the context attributes for a given complex activity
η: all the atomic activities for a given complex activity
μ: all the context attributes for a given complex activity
ϱ: all the $A_{tð}$ for a given complex activity
ω: all the $C_{tð}$ for a given complex activity

Equation (1) models all the possible different ways by which a complex activity may be performed by different users based on their diversity and internal as well as external factors affecting the activity, which could include environment-based distractions, false starts, delayed completion, and failed attempts leading to missing one or more $A_{ts}$, $C_{ts}$, $A_{tE}$, $C_{tE}$, $A_{tð}$, $C_{tð}$, $A_{tI}$, and $C_{tI}$ [82]. Equation (2) models all the possible scenarios by which a complex activity may be performed in which the user, irrespective of their diversity or the effect of internal and external factors, would always reach the end goal or the desired outcome. Equation (3) models all those scenarios of performing a complex activity where the user would never reach the end goal or outcome. All these equations [82] were developed by using probabilistic reasoning principles, concepts from the binomial theorem [83], and by reasoning-based analysis of the dynamic weights for each of $A_{ts}$, $C_{ts}$, $A_{tE}$, $C_{tE}$, $A_{tð}$, $C_{tð}$, $A_{tI}$, and $C_{tI}$, during each complex activity instance [81]. Based on these equations, the work proposed in [82] considers any instance of a complex activity as a designated set that consists of {$A_{ts}$, $C_{ts}$, $A_{tE}$, $C_{tE}$, $A_{tð}$, $C_{tð}$, $A_{tI}$, $C_{tI}$, η, μ, ϱ, ω, ζ(t), Θ(t), Ψ(t)}, where $A_{ts}$, $C_{ts}$, $A_{tE}$, $C_{tE}$, $A_{tð}$, and $C_{tð}$ are determined based on a weighted approach that considers the weights associated to different $A_{tI}$ and $C_{tI}$. Thereafter, η, μ, ϱ, and ω are computed by applying probabilistic reasoning principles to the sequence of $A_{tI}$ and $C_{tI}$ for the given complex activity [81]. Analysis of the different ways by which a typical complex activity of eating lunch may be performed in a typical environment [84] is shown in Table 1.

**Table 1.** Determining the different characteristics of a typical complex activity—eating lunch.

| Characteristics | Feature Description |
|---|---|
| At1 | Standing (0.08) |
| Ct1 | Lights on (0.08) |
| At2 | Walking towards dining table (0.20) |
| Ct2 | Dining area (0.20) |
| At3 | Serving food on a plate (0.25) |
| Ct3 | Food present (0.25) |
| At4 | Washing hand/using hand sanitizer (0.20) |
| Ct4 | Plate present (0.20) |
| At5 | Sitting down (0.08) |
| Ct5 | Sitting options available (0.08) |
| At6 | Starting to eat (0.19) |
| Ct6 | Food quality and taste (0.19) |
| $A_{ts}$ | {At1, At2} |
| $C_{ts}$ | {Ct1, Ct2} |
| $A_{tE}$ | {At5, At6} |
| $C_{tE}$ | {Ct5, Ct6} |
| $A_{tð}$ | {At2, At3, At4} |
| $C_{tð}$ | {Ct2, Ct3, Ct4} |
| $A_{tI}$ | {At1, At2, At3, At4, At5, At6} |
| $C_{tI}$ | {Ct1, Ct2, Ct3, Ct4, Ct5, Ct6} |
| η | 6 |
| μ | 6 |
| ϱ | 4 |
| ω | 4 |
| ζ(t) | 64 |
| Θ(t) | 4 |
| Ψ(t) | 60 |

As can be seen from Table 1, ζ(t): all the different ways by which this complex activity can be performed by different users = 64, Θ(t): all the different ways by which this complex activity can be performed where the specific user performing the activity always reaches the end goal = 4, and Ψ(t): all the different ways of performing this complex activity where the specific user performing the activity never reaches the end goal = 60. Modeling all these instances for different activities in a given environment, as per the methodology discussed in [82], is important towards understanding the specific navigational and movements patterns for determining the indoor location of the specific user as well as other components of the user's indoor position such as building information, floor description, and spatial orientation in terms of whether the user is inside or outside a given indoor spatial region.

It is important that the tracking of the user's indoor location along with these components of their position is highly accurate to improve the trust and acceptance of such AAL technologies as well as to contribute towards improved quality of life and enhanced user experience. Therefore, our framework uses a novel combination of Gradient Boosted Trees along with the AdaBoost [85] approach to achieve greater levels of performance accuracy. Gradient Boosting and the AdaBoost algorithm are two of the most popular boosting approaches that are applied to machine learning systems to bolster their performance accuracies [85]. Gradient Boosted Trees (Gradient Boosting applied to decision trees) have been used by researchers [86,87] in the recent past to achieve high-performance accuracies for activity analysis. Similarly, the AdaBoost algorithm has also been of interest to researchers [88,89] in the field of activity recognition and analysis for optimizing the performance accuracy of the proposed systems. However, a combination of these boosting approaches has not been used in the field of indoor localization thus far. Therefore, we propose this novel approach in our framework that integrates the Gradient Boosted Trees with the AdaBoost algorithm using k-means cross-validation [85] to improve the overall performance accuracy, which also helps to remove false positives and overfitting of data.

Gradient Boosted Trees is a machine learning approach that is used for a range of classification and regression-related tasks where a high-performance accuracy is necessary. The output of this method is an ensemble of decision tree-based predictive models, and it usually outperforms the random forest-based learning approach [90,91]. The methodology builds the learning model in a stage-wise and iterative manner similar to other boosting methods in machine learning, and it generalizes them by allowing optimization of an arbitrary differentiable loss function [92]. The methodology by which the Gradient Boosted Trees approach works is represented by Equations (4)–(8).

$$\hat{F} = arg_F min \mathbb{E}_{x,y}[L(y, F(x))] \tag{4}$$

$$\hat{F}(x) = \sum_{i=1}^{M} \gamma_i h_i(x) + const. \tag{5}$$

$$F_0(x) = arg_\gamma min \sum_{i=1}^{n} L(y_i, \gamma) \tag{6}$$

$$F_m(x) = F_{m-1}(x) - \gamma \sum_{i=1}^{n} \nabla_{F_{m-1}} L(y_i, F_{m-1}(x_i)) \tag{7}$$

$$\gamma_m = arg_\gamma min \sum_{i=1}^{n} L(y_i, F_m)) = arg_\gamma min \sum_{i=1}^{n} L(y_i F_{m-1}(x_i) - \gamma \nabla F_{m-1} L(y_i, F_{m-1}(x_i))) \tag{8}$$

where:
- y = output variable
- x = vector of input variables
- $\hat{F}(x)$ = a function that best approximates the output variable from the input variables
- $F_m$ = some imperfect model
- $L(y, F(x))$ = the loss function
- $h_m$ = the base learner function
- $h_i(x)$ = weighted sum of functions
- $F_0(x)$ = constant function of the model to minimize empirical risk
- $\gamma_m$ = step length

AdaBoost, an abbreviation for Adaptive Boosting [93,94], is a machine learning-based algorithm, which can "boost" or improve the performance accuracy of machine learning-based classification models when used with them in combination. The algorithm uses specific characteristic features from the training data that would help to "boost" the accuracy of the final prediction in the test data. This concept helps to reduce dimensionality and improves the operation time, as those characteristic features of the data that do not significantly contribute towards the classification task are not calculated anymore. The algorithm helps to reduce the overfitting of data as well. After AdaBoost is applied to any machine learning-based classifier, the boosted version of the same can be represented as per Equations (9) and (10).

$$\lambda_N(x) = \sum_{n=1}^{N} \phi_n(x) \tag{9}$$

$$\sigma_t = \sum_m \sigma(\lambda_{n-1}(x_m) + \Delta(n) h(x_m)) \tag{10}$$

where:
- $\phi_n(x)$ = machine learning classifier that uses (x) as input
- $h(x_m)$ = output hypothesis of the machine learning classifier
- $n$ = variable that represents each iteration
- $\sigma_t$ = sum of training error of the final t-stage boosted classification model
- $\lambda_{n-1}(x_m)$ = the boosted classifier upon application of AdaBoost
- $\sigma(f)$ = error function of "f"
- $\phi_n(x) = \Delta(n) h(x_m)$ = the machine learning classifier that is being boosted

The k-folds cross-validation method [95,96] involves an iterative process to train the specific machine learning-based approach for which the k-folds cross-validation is used. The methodology splits the dataset into "k" folds or subsets at the beginning of the process. Thereafter, for each iteration, it uses (k − 1) subsets to train the learner on the remaining subset. As this iterative process runs for "k" number of times, so the methodology is known as k-folds cross-validation. The approach determines the final performance accuracy of the learner by combining the results from all these iterations. This helps to improve the accuracy of the learner. The Mean Squared Error (MSE) is determined by this approach, as shown in Equation (11).

$$MSE = \frac{1}{n} \sum_{i=1}^{n} (y_i - \hat{y}_i)^2 = \frac{1}{n} \sum_{i=1}^{n} (y_i - a - \beta^T x_i)^2 \tag{11}$$

where:
- *MSE* = Mean Squared Error
- $y_i$ = real response values
- $x_i$ = p dimensional vector covariates
- $\hat{y}$ = the hyperplane function

a = coefficient of the first term of the hyperplane function
$\beta^T$ = coefficient of the second term of the hyperplane function
$n$ = number of p dimensional vector covariates

To develop our framework that consisted of the above methodologies, we used RapidMiner [97]. RapidMiner is a software application development platform that allows the development, implementation, and integration of various Big Data, Machine Learning, Artificial Intelligence, Internet of Things, and Information Retrieval-related algorithms. We used RapidMiner specifically for two reasons—(1) the associated functionalities of RapidMiner allow seamless development of an application that allows integration of various methodologies and advancements from all the above-mentioned disciplines of computer science and (2) the characteristic features of RapidMiner allows development and customization of such an application that can communicate with other platforms and can be integrated with the same. We used the RapidMiner Studio with version 9.9.000 installed on a Microsoft Windows 10 computer for the development of this framework. The specifications of the computer were Intel (R) Core (TM) i7-7600U CPU @ 2.80 GHz, two core(s), and four logical processor(s). The educational version of RapidMiner Studio was used to overcome the data processing limit of the free version of this software. As we have used RapidMiner to develop this proposed methodology, we define two terminologies specific to RapidMiner here in a broad manner. These are "process" and "operator." In RapidMiner, an "operator" refers to one of the multiple building blocks of an application associated with specific functionalities that can be changed or modified either statically or dynamically, both by the user and by the application based on the specific need. In RapidMiner, certain "operators" are already developed in the software tool, which can be customized or updated. The tool also allows the development of new "operators" based on any specific need. A continuous, logical, and operational collection of "operators" linearly or hierarchically representing a working application with one or more output characteristics is referred to as a "process" [98].

We have developed a Big Data collection system, consisting of wearables and wireless sensors, in our lab space that can collect the various forms of behavioral, navigational, and localization-related data from real-time experiments involving different users interacting with multimodal contextual parameters to perform different activities in a simulated IoT-based pervasive living environment [99]. To conduct experiments in real-time with human subjects by using this Big Data-based data collection system for the development of the proposed framework, we have obtained Institutional Review Board (IRB) approval [100] from our institution. We also developed the experimental protocols for data collection from the participants and were planning on recruiting them for our study. However, on account of the surge of COVID-19 cases, different sectors of the government in the United States recommended working from home [101]; therefore, we could not conduct real-time data collection experiments using our Big Data collection system. Therefore, we used a dataset for the development and evaluation of our framework. The dataset that was selected was the dataset developed by Torres-Sospedra et al. [102]. This dataset was selected for multiple reasons, which include—(1) the attributes present in this dataset are the same as the attributes of the data that can be collected by using our Big Data collection system; (2) this dataset consists of behavioral and navigational data from multiple different users—which would help our framework to model the characteristics of these individual users to take a personalized approach for detecting their specific location and specific needs; (3) the dataset contains attributes that consist of building, floor, and spatial information which can be used by our framework to provide additional semantic context and meaning to indoor location detections.

The authors [102] developed this dataset by conducting a comprehensive data collection that involved 20 different users who used 25 different android devices and navigated in the 3 buildings of Universitat Jaume I [103], each of which had 4 or more floors, with an approximate area of 110.000 m². There are 529 attributes in the dataset that consists of Wi-Fi fingerprint information, building data, floor information, spatial orientation data— in terms of whether the user was inside or outside a specific indoor region on a specific floor, and the latitude and longitude information of the user's indoor position. The Wi-Fi fingerprint information was collected by the authors from wireless access points (WAPs) and by tracking the RSSI data. The intensity of this data was found to be in the range of -104 dBm to 0 dbM, where the value of -104 dBm represented a very poor signal. The data collection process involved collecting the data from 520 WAP's, and a positive value of 100 was used by the authors to indicate the instance of no detection of a WAP in the user's instantaneous location. Different aspects of the RapidMiner "process" for the development of these multimodal functionalities of our framework by using this dataset [102] are shown in Figures 1–3.

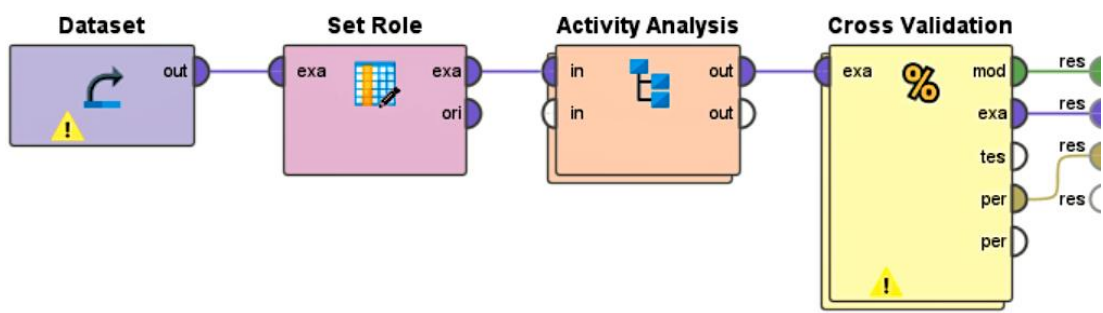

**Figure 1.** The process developed in RapidMiner to develop and implement the proposed framework for personalized indoor localization.

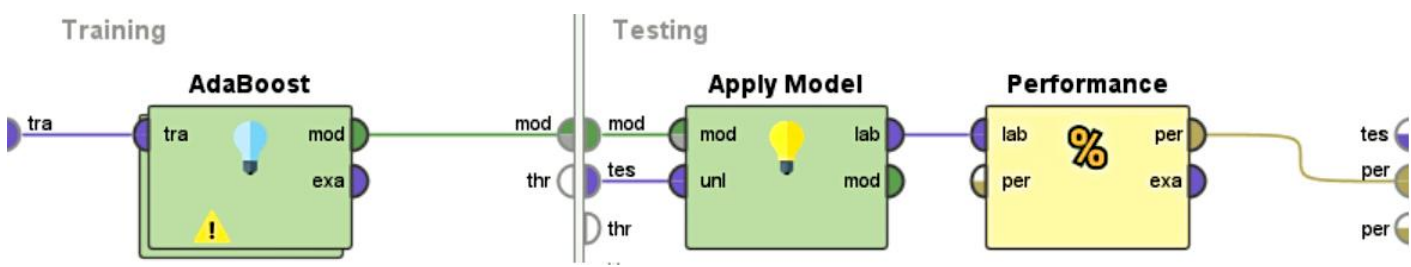

**Figure 2.** The cross-validation sub-process of the process shown in Figure 1 that shows the training and testing components of the main RapidMiner process.

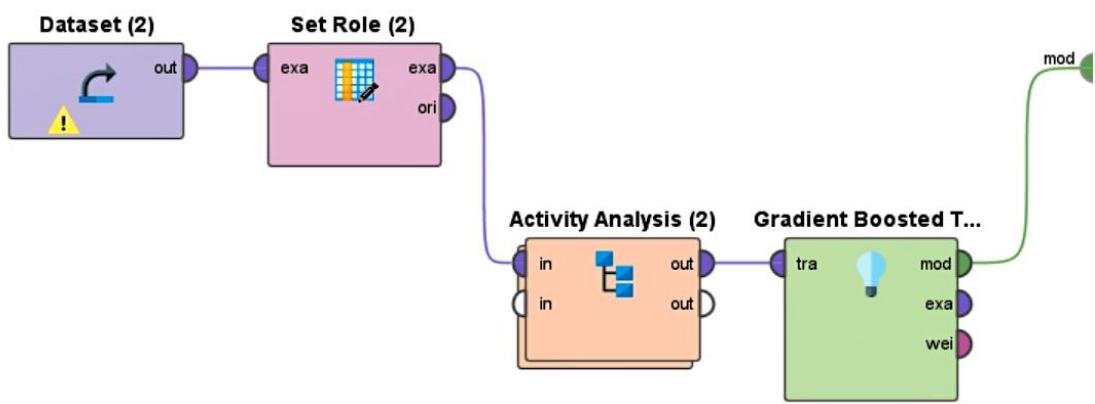

**Figure 3.** The sub-process of the process shown in Figure 2 that shows the development of the AdaBoost algorithm in this framework that used the Gradient Boosted Trees.

We used the 'Dataset' operator in RapidMiner to import this dataset in the RapidMiner software platform, as shown in Figure 1. Then, the 'Set Role' operator was used to instruct the RapidMiner process about the specific attribute that it should predict. We updated the working of this operator on two different occasions to have the RapidMiner process detect the floor and spatial context of the user, respectively. Thereafter, we developed the 'Activity Analysis' operator to implement the activity analysis methodologies in terms of analyzing each complex activity as $A_{ts}$, $C_{ts}$, $A_{tE}$, $C_{tE}$, $A_{tð}$, $C_{tð}$, $A_{tI}$, $C_{tI}$, η, μ, ϱ, ω, ζ(t), Θ(t), Ψ(t), as outlined at the beginning of this section. The objective of this operator was to ensure that the model could detect and analyze all possible forms of user interactions related to any given activity on account of user diversity which would also comprise the specific set of user interactions of any actual user in the real world. Thereafter, we used the cross-validation operator and supplied the value of k=10 for this operator. Figure 2 shows the sub-process that outlines the working of this cross-validation operator. This sub-process consisted of two modules—the training module and the testing module. The training module consisted of the 'AdaBoost' operator that developed and implemented the AdaBoost algorithm to run for a total of 10 iterations. The testing module consisted of the 'Apply Model' operator, which was used to apply the model on the test set and the 'Performance' operator evaluated the performance characteristics of the model for each such iteration. Figure 3 shows the sub-process that outlines the working of the 'AbaBoost' algorithm and each of its iterations. It also shows the 'Gradient Boosted Trees' operator that was used to develop and implement the Gradient Boosted Learning approach with decision trees in our framework. The characteristics of this operator included—number of trees: 50, maximal depth: 5, minimum number of rows: 10, minimum split improvement: 1.0E-5, number of bins: 20, learning rate: 0.01, sample rate: 1.0, and auto distribution. These features of the Gradient Boosted Trees operator, as well as the number of iterations of the AdaBoost learning operator and the cross-validation operator, were selected after trying and evaluating multiple other features to determine those specific set of features that provided the highest accuracy. The 'Performance' operator in RapidMiner calculates the performance of a learning model in terms of overall accuracy, class precision, and class recall values by using a confusion matrix [104], as shown in Equations (12)–(14), respectively. Details about the results and associated discussions that uphold the relevance of our framework are presented in Section 4.

$$\text{Acc}(P, N) = \frac{T(p)+T(n)}{T(p)+T(n)+F(p)+F(n)} \tag{12}$$

$$\text{Pr}(P, N) = \frac{T(p)}{T(p)+F(p)} \tag{13}$$

$$\text{Re}(P, N) = \frac{T(p)}{T(p)+F(n)} \tag{14}$$

where:
Acc(P, N) = overall accuracy of the learning model
Pr(P, N) = class precision value
Re(P, N) = class recall value
$T(p)$ = true positive
$T(n)$ = true negative
$F(p)$ = false positive
$F(n)$ = false negative

## 4. Results and Discussion

This section presents the results of testing the different functionalities of our framework by using the dataset developed by Torres-Sospedra et al. [102]. The dataset consists of different attributes that presents indoor localization, navigational, and behavioral data of 20 different users provided in two different data files. We used one of those files for the development and evaluation of our framework. This data file that we used consisted of the data of 18 users. This data file had a total of 529 attributes and 19,937 rows. Each row in this data file represents a specific user's localization, navigational, and behavioral data in terms of the RSSI data collected from each of the 520 WAP's from 3 different buildings, consisting of 254 spatial regions, and indicating whether the user was inside or outside these spatial regions, along with the user's latitude and longitude information. We converted some of these attributes from numerical type to polynomial type for the development of our framework as per the proposed system architecture discussed in Section 3. Table 2 outlines the description of these respective attributes present in the dataset.

**Table 2.** Outline of the attributes in the data that were used to develop, implement, and evaluate the proposed framework.

| Attribute Name | Description |
| --- | --- |
| WAP001 | Intensity of Signal obtained from WAP #001 |
| WAP002 | Intensity of Signal obtained from WAP #002 |
| WAP003 | Intensity of Signal obtained from WAP #003 |
| WAP004 | Intensity of Signal obtained from WAP #004 |
| ⋮ | ⋮ |
| WAP520 | The intensity of Signal obtained from WAP #520 |
| Longitude | The longitude of the user's indoor position |
| Latitude | The latitude of the user's indoor position |
| Floor | The specific floor number where the user was located |
| Building | The specific building number where the user was located |
| Space ID | The identifier representing a specific spatial region |
| RelativePosition | States whether the user was inside or outside a specific spatial region |
| User ID | A unique identifier to identify each user |
| Phone ID | The identifier representing the specific phone that was carried by the user |
| Timestamp | The timestamp information associated with the user's location |

As per the system architecture of our framework, as presented and discussed in Section 3, we analyzed the diverse ways in which the localization, navigational, and behavioral data of the 18 different users varied with respect to their instantaneous and dynamic locations in different floors of the different buildings where the data was collected. This analysis is shown in Figures 4–7, respectively.

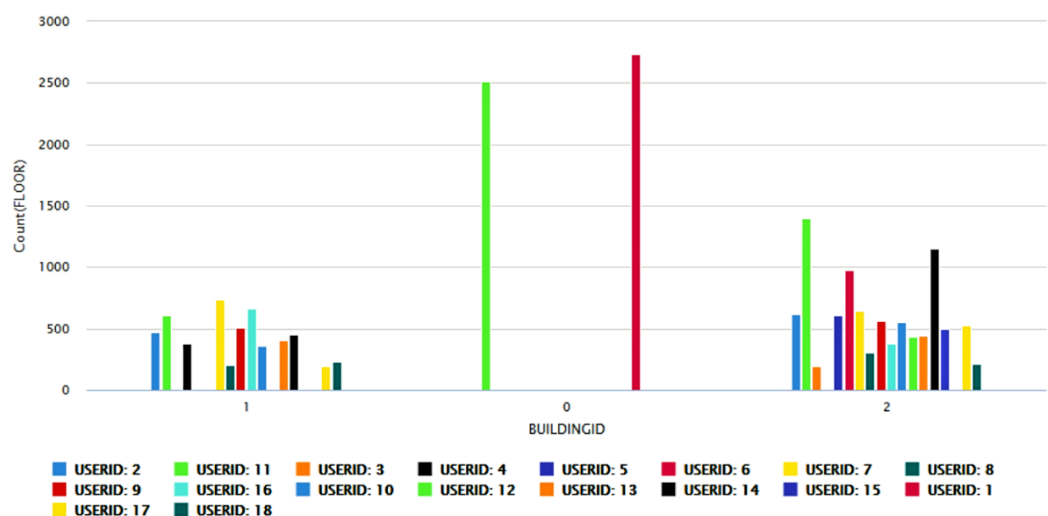

**Figure 4.** A bar plot representing the number of times the different users were present on a floor (all the floors taken together) in the different buildings, where the plot is grouped by the value of the attribute—BuildingID.

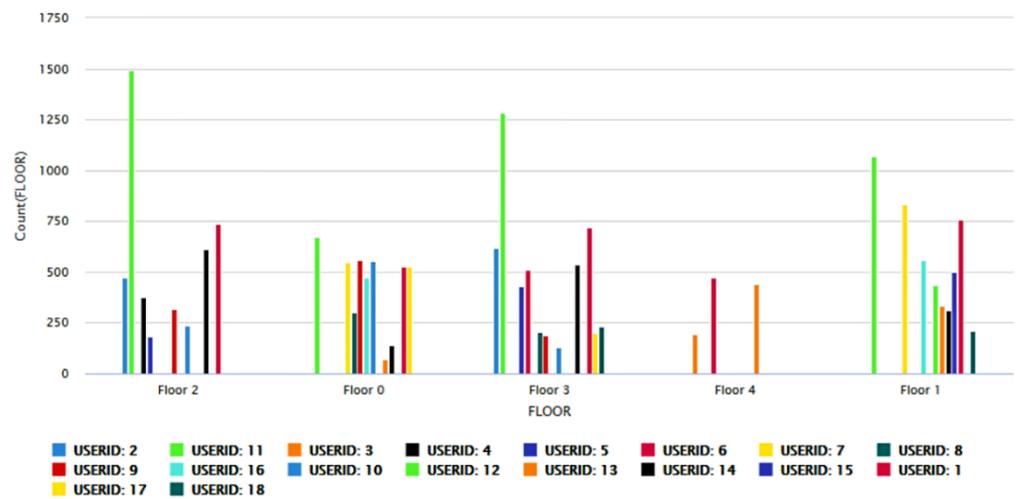

**Figure 5.** A bar plot representing the number of times the different users were present on specific floors in different buildings, where the plot is grouped by the value of the attribute—Floor.

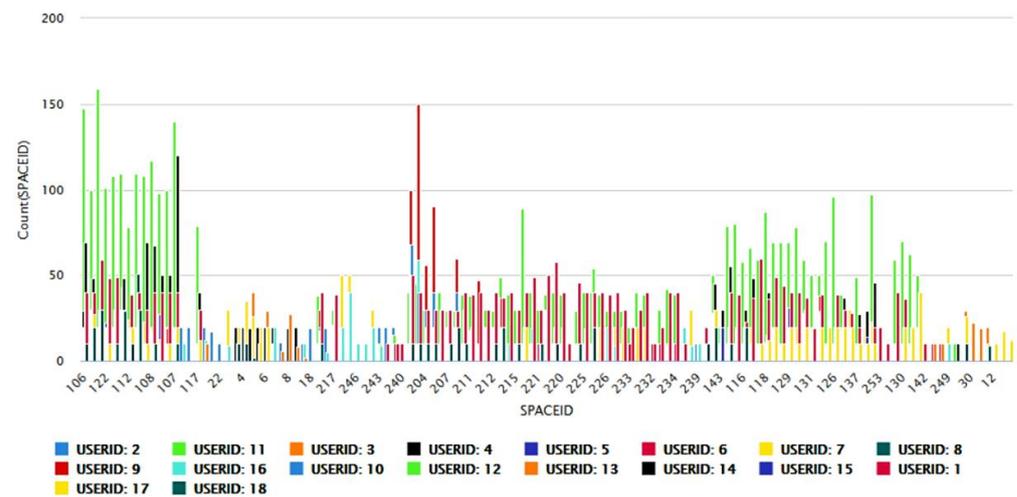

**Figure 6.** A bar plot representing the number of times the different users were present in different spatial regions (not all of them are shown to maintain clarity) in different buildings, where the plot is grouped by the value of the attribute—SpaceID.

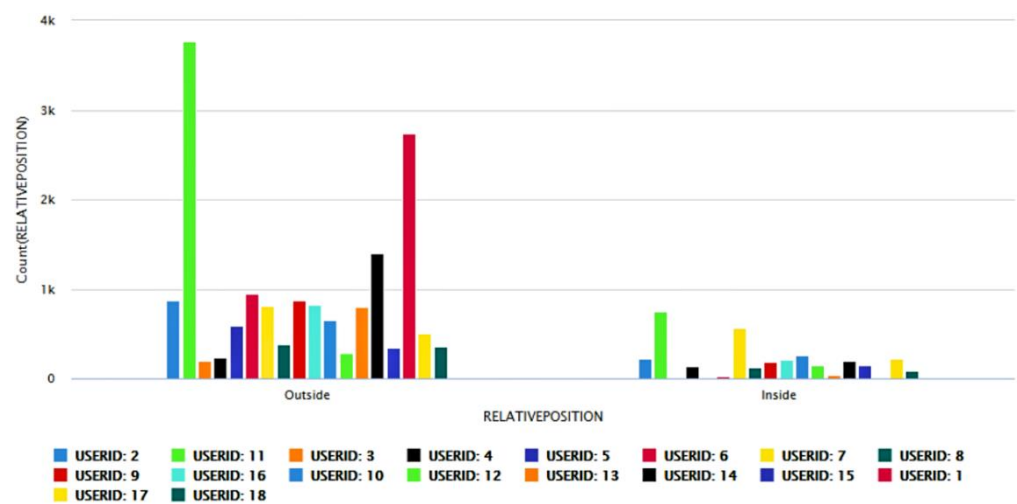

**Figure 7.** A bar plot representing the number of times the different users were either inside or outside a spatial region across the different floors in the different buildings, where the plot is grouped by the value of the attribute—RelativePosition.

After performing this analysis, we developed the functionality in our framework to model individual user profiles by filtering, interpreting, and analyzing the specific set of behavioral, navigational, and location-based information during different activities that were specific to each user, out of all the available data that represented the diverse set of user interactions as per Equations (1)–(3). This was implemented as per the methodology discussed in Section 3, and thereafter, we trained the k-folds cross validation-based machine learning approach that uses the AdaBoost algorithm and the Gradient Boosted Trees (Figures 1–3) to calculate different components of each user's location information during each iteration. In other words, this machine learning approach was run 18 times to model 18 different users by developing individual user profiles of each user, as per the user interactions specific to each to them. We ran this machine learning model another time to calculate the different components of indoor location by considering the data from all the users, or in other words, by following the traditional approach of modeling an average user. This was done to ensure that we have grounds for comparison to evaluate if our approach of modeling specific users as per user interactions specific to each user helped us to achieve higher performance accuracies as compared to the traditional approach of modeling an average user. Every time the machine learning approach was run, we calculated these components of the specific user's indoor location—(1) the floor in a specific building where the specific user was located at a specific time instant and (2) whether the user was located inside or outside a specific indoor spatial region, by taking into consideration the WAP data from all the 254 spatial regions at that time instant. To evaluate the performance of our machine learning approach for each such iteration, we used the confusion matrix (Equations (12)–(14)) to calculate the overall accuracy, class precision values, and class recall values. The results of 19 iterations of this machine learning approach (18 iterations for 18 users + 1 iteration for modeling an average user) for indoor localization in terms of indoor floor detection are summarized and shown in Table 3, where all these performance characteristics have been represented as percentages.

**Table 3.** Summary of results for performing indoor localization in terms of indoor floor detection by modeling specific users and their associated user interactions.

| User | Overall Accuracy | CP Floor 0 | CR Floor 0 | CP Floor 1 | CR Floor 1 | CP Floor 2 | CR Floor 2 | CP Floor 3 | CR F Floor 3 | CP Floor 4 | CR Floor 4 |
|---|---|---|---|---|---|---|---|---|---|---|---|
| Average User | 89.16% | 89.71% | 94.94% | 94.89% | 89.02% | 84.06% | 78.58% | 85.40% | 91.18% | 99.91% | 99.91% |
| User 1 | 95.69% | 99.05% | 98.48% | 98.53% | 97.49% | 98.18% | 88.54% | 94.56% | 99.03% | 0.00% | 0.00% |
| User 2 | 100.00% | 0.00% | 0.00% | 0.00% | 0.00% | 100.00% | 100.00% | 100.00% | 100.00% | 0.00% | 0.00% |
| User 3 | 100.00% | 0.00% | 0.00% | 0.00% | 0.00% | 0.00% | 0.00% | 0.00% | 0.00% | 100.00% | 100.00% |
| User 4 | 100.00% | 0.00% | 0.00% | 0.00% | 0.00% | 100.00% | 100.00% | 0.00% | 0.00% | 0.00% | 0.00% |
| User 5 | 100.00% | 0.00% | 0.00% | 0.00% | 0.00% | 100.00% | 100.00% | 100.00% | 100.00% | 0.00% | 0.00% |
| User 6 | 100.00% | 0.00% | 0.00% | 0.00% | 0.00% | 0.00% | 0.00% | 100.00% | 100.00% | 100.00% | 100.00% |
| User 7 | 98.12% | 96.80% | 99.64% | 99.75% | 97.12% | 0.00% | 0.00% | 0.00% | 0.00% | 0.00% | 0.00% |
| User 8 | 100.00% | 100.00% | 100.00% | 0.00% | 0.00% | 0.00% | 0.00% | 100.00% | 100.00% | 0.00% | 0.00% |
| User 9 | 99.44% | 100.00% | 100.00% | 0.00% | 0.00% | 98.74% | 99.37% | 98.94% | 97.89% | 0.00% | 0.00% |
| User 10 | 100.00% | 100.00% | 100.00% | 0.00% | 0.00% | 100.00% | 100.00% | 100.00% | 100.00% | 0.00% | 0.00% |
| User 11 | 94.13% | 92.60% | 97.16% | 97.96% | 89.87% | 96.90% | 93.98% | 96.79% | 96.26% | 0.00% | 0.00% |
| User 12 | 100.00% | 0.00% | 0.00% | 100.00% | 100.00% | 0.00% | 0.00% | 0.00% | 0.00% | 0.00% | 0.00% |
| User 13 | 100.00% | 100.00% | 100.00% | 100.00% | 100.00% | 0.00% | 0.00% | 0.00% | 0.00% | 100.00% | 100.00% |
| User 14 | 99.37% | 97.84% | 98.55% | 99.35% | 99.03% | 100.00% | 99.34% | 99.45% | 99.81% | 0.00% | 0.00% |
| User 15 | 100.00% | 0.00% | 0.00% | 100.00% | 100.00% | 0.00% | 0.00% | 0.00% | 0.00% | 0.00% | 0.00% |
| User 16 | 98.06% | 98.71% | 97.03% | 97.54% | 98.93% | 0.00% | 0.00% | 0.00% | 0.00% | 0.00% | 0.00% |
| User 17 | 100.00% | 100.00% | 100.00% | 0.00% | 0.00% | 0.00% | 0.00% | 100.00% | 100.00% | 0.00% | 0.00% |
| User 18 | 100.00% | 0.00% | 0.00% | 100.00% | 100.00% | 0.00% | 0.00% | 100.00% | 100.00% | 0.00% | 0.00% |

In Table 3, CP stands for Class Precision, and CR stands for Class Recall. As can be seen from the table, the first row presents the performance characteristics in terms of overall accuracy, class precision, and class recall values for modeling an average user. The rest of the rows outline these characteristics for each specific user by taking into consideration the specific set of user interactions consisting of behavioral, navigational, and location-based information, that were specific to each of them. If we consider a function P, such that P(user) represents the performance characteristics of detecting some component of that user's indoor location, then from Table 3, the following can be concluded by the analysis of these respective performance characteristics:

1. In terms of Overall Accuracy of detecting the different floors: P(User 2) = P(User 3) = P(User 4) = P(User 5) = P(User 6) = P(User 8) = P(User 10) = P(User 12) = P(User 13) = P(User 15) = P(User 17) = P(User 18) > P(User 9) > P(User 14) > P(User 7) > P(User 16) > P(User 1) > P(User 11) > P(Average User).
2. In terms of the Class Precision for detection of the users' location's on Floor 0: P(User 8) = P(User 10) = P(User 13) = P(User 17) > P(User 9) > P(User 1) > P(User 16) > P(User 14) > P(User 7) > P(User 11) > P(Average User). Here, the performance characteristics of User 2, User 3, User 4, User 5, User 6, User 12, User 15, and User 18 were not included in the analysis as those users were never present on Floor 0.
3. In terms of the Class Recall for detection of the users' location's on Floor 0: P(User 8) = P(User 10) = P(User 13) = P(User 17) = P(User 9) > P(User 7) > P(User 14) > P(User 1) > P(User 11) > P(User 16) > P(Average User). Here, the performance characteristics of User 2, User 3, User 4, User 5, User 6, User 12, User 15, and User 18 were not included in the analysis as those users were never present on Floor 0.
4. In terms of the Class Precision for detection of the users' location's on Floor 1: P(User 13) = P(User 12) = P(User 15) = P(User 18) > P(User 7) > P(User 14) > P(User 1) > P(User 11) > P(User 16) > P(Average User). Here the performance characteristics of User 2, User 3, User 4, User 5, User 6, User 8, User 9, User 10, and User 17 were not included in the analysis as those users were never present on Floor 1.
5. In terms of the Class Recall for detection of the users' location's on Floor 1: P(User 13) = P(User 12) = P(User 15) = P(User 18) > P(User 14) > P(User 16) > P(User 1) > P(User 7) > P(User 11) > P(Average User). Here the performance characteristics of User 2, User 3, User 4, User 5, User 6, User 8, User 9, User 10, and User 17 were not included in the analysis as those users were never present on Floor 1.
6. In terms of the Class Precision for detection of the users' location's on Floor 2: P(User 14) = P(User 10) = P(User 2) = P(User 4) = P(User 5) > P(User 9) > P(User 1) > P(User 11) > P(Average User). Here, the performance characteristics of User 3, User 6, User 7, User 8, User 12, User 13, User 15, User 16, User 17, and User 18 were not included in the analysis as those users were never present on Floor 2.
7. In terms of the Class Recall for detection of the users' location's on Floor 2: P(User 10) = P(User 2) = P(User 4) > P(User 5) > P(User 9) > P(User 14) > P(User 11) > P(User 1) > P(Average User). Here, the performance characteristics of User 3, User 6, User 7, User 8, User 12, User 13, User 15, User 16, User 17, and User 18 were not included in the analysis as those users were never present on Floor 2.
8. In terms of the Class Precision for detection of the users' locations on Floor 3: P(User 10) = P(User 2) = P(User 5) = P(User 18) = P(User 8) = P(User 17) = P(User 6) > P(User 14) > P(User 9) > P(User 11) > P(User 1) > P(Average User). Here, the performance characteristics of User 3, User 4, User 7, User 12, User 13, User 15, and User 16 were not included in the analysis as those users were never present on Floor 3.
9. In terms of the Class Recall for detection of the users' locations on Floor 3: P(User 10) = P(User 2) = P(User 5) = P(User 18) = P(User 8) = P(User 17) = P(User 6) > P(User 14) > P(User 1) > P(User 9) > P(User 11) > P(Average User). Here, the performance characteristics of User 3, User 4, User 7, User 12, User 13, User 15, and User 16 were not included in the analysis as those users were never present on Floor 3.

10. In terms of the Class Precision for detection of the users' locations on Floor 4: P(User 6) = P(User 13) = P(User 3) > Average User. Here, the performance characteristics of the other users were not included because other than User 6, User 13, and User 3, none of the other users were present on Floor 4.
11. In terms of the Class Recall for detection of the users' locations on Floor 4: (User 6) = P(User 13) = P(User 3) > Average User. Here, the performance characteristics of the other users were not included because other than User 6, User 13, and User 3, none of the other users were present on Floor 4.

Based on this analysis and the findings of the above methodology for personalized indoor localization with a specific focus on detecting the floor where a specific user was located, that modeled the dynamic behavioral, navigational, and localization patterns of 18 different users across 3 different buildings consisting of 5 different floors, it can be concluded that this approach is always superior to the traditional approach of modeling an average user as the performance characteristics (overall accuracy, class precision, and class recall) for any specific user (obtained by taking into consideration the user interactions specific to that user) are always higher than the performance characteristics of the average user.

Thereafter, we modeled the behavioral, navigational, and localization patterns of all these specific users to interpret details of their spatial location to determine if they were located inside or outside a given indoor spatial region. The data consisted of 254 distinct and non-overlapping regions where one or more users were present at different times, and the data obtained from the WAP's present in these zones were considered for development of this approach as discussed in Section 3. For each such user, our machine learning-based model provided performance characteristics in terms of overall accuracy, class precision, and class recall associated with the detection of whether that specific user was located inside, or outside a given spatial region. We also modeled the characteristics of an average user for grounds of comparison. These results are compiled and shown in Table 4, where CP stands for class precision, and CR stands for class recall.

**Table 4.** Summary of results for performing indoor localization to detect whether a specific user is located inside or outside a specific indoor spatial region out of the 254 indoor spatial regions.

| User | Overall Accuracy | CP Inside | CR Inside | CP Outside | CR Outside |
|---|---|---|---|---|---|
| Average User | 77.17% | 40.05% | 73.96% | 93.71% | 77.81% |
| User 1 | 100.00% | 0.00% | 0.00% | 100.00% | 100.00% |
| User 2 | 91.29% | 80.66% | 76.00% | 93.86% | 95.27% |
| User 3 | 100.00% | 0.00% | 0.00% | 100.00% | 100.00% |
| User 4 | 96.51% | 98.46% | 92.09% | 95.49% | 99.15% |
| User 5 | 99.51% | 100.00% | 82.35% | 99.50% | 100.00% |
| User 6 | 99.69% | 100.00% | 90.00% | 99.69% | 100.00% |
| User 7 | 88.14% | 78.48% | 98.07% | 98.36% | 81.20% |
| User 8 | 94.29% | 87.70% | 88.43% | 96.36% | 96.11% |
| User 9 | 96.71% | 98.12% | 83.07% | 96.47% | 99.66% |
| User 10 | 95.72% | 93.98% | 90.70% | 96.39% | 97.71% |
| User 11 | 95.59% | 90.50% | 81.88% | 96.49% | 98.30% |
| User 12 | 96.56% | 94.12% | 96.00% | 97.89% | 96.86% |
| User 13 | 100.00% | 100.00% | 100.00% | 100.00% | 100.00% |
| User 14 | 99.00% | 99.44% | 92.19% | 98.94% | 99.93% |
| User 15 | 92.58% | 93.23% | 81.58% | 92.33% | 97.40% |
| User 16 | 96.32% | 98.29% | 83.09% | 95.92% | 99.64% |
| User 17 | 94.47% | 87.98% | 94.47% | 97.56% | 94.48% |
| User 18 | 97.05% | 93.51% | 90.00% | 97.80% | 98.61% |

In Table 4, each row contains the performance characteristics of a specific user in terms of the overall accuracy, class precision, and class recall values of our proposed machine learning-based model to determine whether the user was located inside or outside an indoor spatial region. If we consider a function P, such that P(user) represents the performance characteristics of detecting some component of that user's indoor location, then from Table 4, the following can be concluded by the analysis of these respective performance characteristics:

1. In terms of overall performance accuracy: P(User 1) = P(User 3) = P(User 13) > P(User 6) > P(User 5) > P(User 14) > P(User 18) > P(User 9) > P(User 12) > P(User 4) > P(User 16) > P(User 10) > P(User 11) > P(User 17) > P(User 8) > P(User 15) > P(User 2) > P(User 7) > P(Average User)
2. In terms of class precision for detecting a user inside a given spatial region: P(User 13) = P(User 6) = P(User 5) > P(User 14) > P(User 4) > P(User 16) > P(User 9) > P(User 12) > P(User 10) > P(User 18) > P(User 15) > P(User 11) > P(User 17) > P(User 8) > P(User 2) > P(User 7) > P(Average User). Here the User 1 and User 3 were considered in the analysis as they were never present inside the confines of a spatial region.
3. In terms of class recall for detecting a user inside a given spatial region: P(User 13) > P(User 7) > P(User 12) > P(User 17) > P(User 14) > P(User 4) > P(User 10) > P(User 6) > P(User 18) > P(User 8) > P(User 16) > P(User 9) > P(User 5) > P(User 11) > P(User 15) > P(User 2) > P(Average User). Here the User 1 and User 3 were considered in the analysis as they were never present inside the confines of a spatial region.
4. In terms of class precision for detecting a user outside a given spatial region: P(User 13) > P(User 1) > P(User 3) > P(User 6) > P(User 5) > P(User 14) > P(User 7) > P(User 12) > P(User 18) > P(User 17) > P(User 11) > P(User 9) > P(User 10) > P(User 8) > P(User 16) > P(User 4) > P(User 2) > P(Average User) > P(User 15). Here even though P(Average User) was greater than P(User 15), the difference between these values was 0.0138, which was not significant.
5. In terms of class precision for detecting a user outside a given spatial region: P(User 13) = P(User 1) = P(User 3) = P(User 6) = P(User 5) > P(User 14) > P(User 9) > P(User 16) > P(User 4) > P(User 18) > P(User 11) > P(User 10) > P(User 15) > P(User 12) > P(User 8) > P(User 2) > P(User 17) > P(User 7) > P(Average User)

Based on this analysis and the findings of the above methodology for personalized indoor localization to determine if a specific user was located inside or outside an indoor spatial region, that modeled the dynamic behavioral, navigational, and localization patterns of 18 different users across 3 different buildings consisting of 5 different floors and 254 spatial regions, it can be concluded that this approach is always superior to the traditional approach of modeling an average user as the performance characteristics (overall accuracy, class precision, and class recall) for any specific user (obtained by taking into consideration the user interactions specific to that user) are always higher than the performance characteristics of the average user.

## 5. Comparative Discussion

Despite many works [33–73] in the field of indoor localization for AAL, several research challenges exist, as discussed in Section 2. The proposed multifunctional framework for indoor localization for personalized AAL that can model specific users by taking into consideration the diverse and dynamic behavioral, navigational, and localization-related components of user interactions that are specific to each user in a multi-user environment, aims to address these research challenges. In this section, we further discuss how the functionalities, specifications, and performance characteristics of our framework help to address these challenges and outperform similar works in this field. This discussion is presented as follows:

1. Researchers [54–69] in this field have focused on defining an average user in terms of certain attributes such as cognitive, behavioral, perceptual, and mental abilities and then developing indoor localization-based AAL systems to meet the needs of the average user. In the real world, no such average user exists. The user interaction patterns in terms of behavioral, navigational, and localization-related characteristics of each user are different and determined by the diverse characteristic traits specific to a given user, which could include different levels of cognitive, behavioral, perceptual, and mental abilities, just to name a few. Due to this difference in needs and abilities of each specific user, they quite often do not 'fit' into the definition of an average user for whom an AAL system is developed, which results in the ineffectiveness or failure of the associated AAL system to address the needs of a specific user. It is crucial for the future of AAL systems to have a "personalized" touch so that such systems can cater to the dynamic and diverse needs of each specific user. Our framework addresses this challenge in multiple ways. First, it presents a probabilistic reasoning-based mathematical approach (Equations (1)-(3)) to model all the diverse ways by which any given activity can be performed by different users based on internal factors such as physical, mental, cognitive, psychological, and emotional factors, and external factors such as environment variables and context attributes [28–30], that are specific to each user. This analysis is done by breaking down the activity into fine-grain components—atomic activities, context attributes, core atomic activities, core context attributes, other atomic activities, other context attributes, start atomic activities, end atomic activities, start context attributes, and end context attributes (Table 1). Second, our framework consists of the methodology (Section 3) to model multimodal components of the indoor location of each specific user by modeling every user in terms of their distinct behavioral, navigational, and localization-related characteristics during different activities. Upon testing our approach on a dataset that consisted of 18 different users, each of whom exhibited different behavioral, navigational, and localization-related characteristics during different activities we observed that the performance characteristics (in terms of overall accuracy, class precision, and class recall values) of modeling each specific user are always higher than the traditional approach of modeling an average user (Table 3 and 4).

2. Prior works [33–48] in this field that used different forms of machine learning and artificial intelligence approaches to detect the indoor location of a user have used some of the major machine learning approaches without any attempts to boost the performance accuracies of the underlining systems and applications. For the seamless acceptance of the future of AAL technologies that can adapt with respect to the diverse needs of different users, it is crucial to investigate approaches for improving the performance accuracies of such AAL systems. Gradient Boosting and AdaBoost learning approaches are two amongst the most popular methodologies for boosting the performance accuracies of machine learning systems while removing overfitting of data and false positives [85]. Both the Gradient Boosted approach and the AdaBoost approach have achieved promising results in the field of activity recognition for boosting the performance characteristics of machine learning-based activity recognition systems on which they were applied [86–89]. However, these boosting approaches have not been investigated for indoor localization. Moreover, a combination of these two boosting approaches to achieve even higher performance accuracies has not been investigated before in this field of research. Therefore, our framework implements these two boosting approaches together on the decision tree classifier for modeling specific users to detect multimodal components of their indoor location, which includes detecting the floor the specific user is located on (Table 3) and tracking whether the specific user is located inside or outside a given indoor spatial region (Table 4) at a given point of time. Table 5 shows a comparison of different works in the field of indoor localization that used machine learning systems to further uphold the fact that our framework is the first work in this field that used a combination of two boosting approaches to achieve high-performance accuracies while modeling specific users as per their diverse characteristics leading to varying user interactions.

**Table 5.** Comparison of our proposed framework with other similar works in this field that used machine learning approaches for indoor localization.

| Work(s) | Machine Learning Approach | Gradient Boosting | AdaBoost |
|---|---|---|---|
| Varma et al. [33], Gao et al. [34] | Random Forest | - | - |
| Khan et al. [35], Labinghisa et al. [36], Qin et al. [37] | Artificial Neural Network | - | - |
| Musa et al. [38], Yim et al. [39] | Decision Tree | - | - |
| Sjoberg et al. [40], Zhang et al. [41] | Support Vector Machine | - | - |
| Zhang et al. [42], Ge et al. [43], Hu et al. [44] | k-NN | - | - |
| Zhang et al. [45], Poulose et al. [46] | Deep Learning | - | - |
| Jamâa et al. [47], Barsocchi et al. [48] | Linear Regression | - | - |
| Thakur et al. [this work] | Decision Tree | ✓ | ✓ |

3. Research works [54–69] in this field that used the data from multiple users to train the underlining machine learning systems did not have a significant number of participants or volunteers to represent the diversity of actual users. In view of the diversity of the elderly and their varying associated needs, both temporary and permanent, on account of their declining abilities of different degrees, it is crucial that such AAL-based machine learning systems are trained with sufficient data from different users so that the underlining systems are familiar with the user diversity and can achieve high levels of performance accuracy while detecting the location-related information of specific users. Table 6 shows the comparison of our framework with similar works in this field that used the data from one or more users for proposing indoor localization systems. As can be seen from Table 6, our framework uses the maximum number of users to train the boosted learning approach (Section 3) with an aim to train the learning model on diverse user interaction patterns arising from different users while being able to model each of these users by taking into consideration the specific characteristics of their behavioral, navigational, and localization-related information.

**Table 6.** Comparison of our proposed framework with other similar works in this field in terms of the number of users whose user interaction data were used to train the associated machine learning systems.

| Work(s) | Number of Users |
|---|---|
| Xu et al. [63] | 2 |
| Qian et al. [57] | 3 |
| Fusco et al. [58] | 3 |
| Chang et al. [59] | 3 |
| Wang et al. [64] | 3 |
| Kothari et al. [56] | 4 |
| Subbu et al. [60] | 4 |
| Röbesaat et al. [65] | 4 |
| Wu et al. [67] | 4 |
| Chen et al. [62] | 6 |
| Gu et al. [68] | 8 |
| Zhou et al. [61] | 10 |
| Murata et al. [54] | 10 |
| Yoo et al. [55] | 10 |
| Yang et al. [66] | 10 |
| Niu et al. [69] | 15 |
| Thakur et al. [this work] | 18 |

4. The indoor localization-related works in this field have mostly focused on detecting the X and Y coordinate of the user's position. The X and Y coordinate information of a user's indoor position are important attributes of the location information. However, in a real-world scenario where the user could be located in certain spatial regions, such as an apartment which could be located on a specific floor inside a multistoried building, just the X and Y coordinates do not provide enough context as far as the user's location is concerned. In other words, it is not possible to detect the floor or spatial information (such as inside or outside a given indoor spatial region) of a user's location just based on the interpretation of the X and Y coordinate information. This lack of semantic context is likely to lead to delay of care for the elderly for emergency needs such as unconsciousness from a fall, as the emergency responders or healthcare providers would have to resort to a trial-and-error approach until they arrive at the specific floor and in the specific spatial region where the elderly might be located in the multistoried building. Our framework addresses this challenge by being able to detect the floor information (Section 3, Table 3) as well as the dynamic spatial information of the user in terms of whether the user is located inside or outside a given spatial region which is located indoors (Section 3, Table 4). With the methodology to model individual user profiles for personalized indoor localization, our framework achieves high accuracies for floor detection as well as for indoor spatial region detection by using a novel methodology that involved the integration of two boosting approaches. Upon testing of our framework by using a dataset that consisted of the data of 18 different users, each of whom exhibited different behavioral, navigational, and localization-related characteristics during different activities, performed in 3 buildings consisting of 5 floors and 254 spatial regions; we observed that for multiple users our framework achieved 100% accuracy both for floor detection and for spatial information detection. Table 7 shows how this functionality of spatial information detection in terms of detecting whether a user is present inside or outside the confines of an indoor spatial region addresses the limitations in similar works [49–69] in this field in terms of functionality and operational features.

**Table 7.** Comparison of the functionality of indoor spatial information detection in our proposed framework with other similar works in this field.

| Work(s) | Indoor Location Detection | Indoor Spatial Information |
|---|---|---|
| Bolic et al. [49] | ✓ | - |
| Angermann et al. [50] | ✓ | - |
| Evennou et al. [51] | ✓ | - |
| Wang et al. [52] | ✓ | - |
| Klingbeil et al. [53] | ✓ | - |
| Xu et al. [63] | ✓ | - |
| Qian et al. [57] | ✓ | - |
| Fusco et al. [58] | ✓ | - |
| Chang et al. [59] | ✓ | - |
| Wang et al. [64] | ✓ | - |
| Kothari et al. [56] | ✓ | - |
| Subbu et al. [60] | ✓ | - |
| Röbesaat et al. [65] | ✓ | - |
| Wu et al. [67] | ✓ | - |
| Chen et al. [62] | ✓ | - |
| Gu et al. [68] | ✓ | - |
| Zhou et al. [61] | ✓ | - |
| Murata et al. [54] | ✓ | - |
| Yoo et al. [55] | ✓ | - |
| Yang et al. [66] | ✓ | - |
| Niu et al. [69] | ✓ | - |
| Thakur et al. [this work] | ✓ | ✓ |

While there have been a few works [74–80] in indoor floor detection in the recent past, the underlining systems are not highly accurate to support their widescale deployment and real time implementation. To contribute towards increased trust in technology and seamless integration of such AAL systems, it is crucial that the future of indoor floor detection systems consist of the functionality to detect the floor-level information of the user's indoor position in a highly accurate manner while removing false positives and overfitting of data. By implementing a novel approach that involves the use of two boosting algorithms—Gradient Boosting and the AdaBoost algorithm [85] via the use of the k-folds cross-validation, our framework addresses these issues of false positives and overfitting of data while being able to detect the floor-level information of the user's indoor position with a high level of accuracy. Table 8 shows the comparison of the performance characteristics of the floor detection functionality of our framework with these recent works that outline how our framework outperforms all these recent works in this field of research. In Table 8, we list the performance accuracy of our framework for floor detection as 100% because it achieved 100% accuracy for multiple users, as presented in Table 3.

**Table 8.** Comparison of the indoor floor detection functionality in our framework with recent works in indoor floor detection.

| Work(s) | Accuracy |
|:---:|:---:|
| Alsehly et al. [74] | 81.3% |
| Garcia et al. [78] | 81.8% |
| Delahoz et al. [80] | 82.0% |
| Ruan et al. [79] | 85.6% |
| Campos et al. [75] | 90.6% |
| Sun et al. [76] | 93.7% |
| Haque et al. [77] | 93.8% |
| Thakur et al. [this work] | 100.0% |

As per our best knowledge, no similar dataset exists that consists of the localization-related Big Data of 18 or more users recorded during different activities in an IoT-based environment in a user-specific manner, therefore the proposed framework could not be tested on any other dataset other than the one [102] discussed in Section 4. To add, in addition to the nature of comparison of the performance characteristics and operational features of the proposed framework with prior works in this field (Section 2), as discussed in this section, the results could also have been possibly compared by simulating the exact conditions and settings of these prior works and evaluating the framework in those settings. However, that would have involved including volunteers or participants in the study for real-time Big Data collection. This could not be performed on account of the stay-at-home and remote work guidelines recommended by various sectors of the government in the United States on account of the surge in COVID-19 cases [101,105,106] at the time this research was conducted. Upon relaxation of these stay-at-home and remote work guidelines, hopefully soon, we aim to address these limitations by conducting real-time experiments with human subjects in simulated IoT-based environments.

## 6. Conclusions and Scope for Future Work

It is the need of the hour to develop AAL systems that can take a personalized approach to adapt, respond, anticipate, and address the needs of specific users, especially the elderly, to contribute towards their healthy aging and independent living in the future of smart environments, as the constantly increasing population of older adults is demonstrating wide ranging diverse and dynamic needs, that the traditional AAL systems with methodologies to assist an average user are failing to address. Tracking and studying the behavioral, navigational, and localization-related characteristics, distinct to each user in a given indoor environment consisting of multiple users, through user-specific indoor localization during different activities, holds the potential towards addressing the challenges for the creation of personalized AAL living experiences in the future of smart living spaces, such as Smart Homes. Therefore, this paper proposes a multifunctional framework for personalized AAL of multiple users through the lens of indoor localization. The paper makes multiple scientific contributions to this field with an aim to address these challenges as well as the limitations and drawbacks in the existing works in this field.

First, to address the fact that an average user for whom smart systems have been traditionally developed does not exist in the real world, the framework presents two novel methodologies. The first methodology is a probabilistic reasoning-based mathematical approach to model all possible range of user interactions that can be associated with an activity by taking into consideration user diversity and the associated variations in user interactions of each specific user in a multi-user-based smart environment. Thereafter, it proposes a machine learning-based methodology that uses this probabilistic reasoning-based mathematical approach to model individual user profiles to study, anticipate, interpret, and analyze the behavioral, navigational, and localization-related characteristics of each specific user during different activities to detect their specific locations in an indoor environment at different time instants. This methodology was tested on a dataset that consisted of the user interaction data of 18 users during different activities. The results presented and discussed uphold the fact that modeling individual users through user personalization always helps to achieve higher levels of performance accuracy for indoor localization (in terms of overall accuracy, class precision, and class recall values) as compared to the traditional approach of modeling an average user.

Second, to address the need for the development of highly accurate indoor localization systems, the framework presents a novel boosting methodology by integrating two commonly used boosting approaches—the Gradient Boosting approach and the AdaBoost algorithm with the k-folds cross-validation method, for the development of user-specific indoor localization that users of such systems can trust and seamlessly accept into their lives on account of the superior performance accuracies related to detecting multimodal components of their indoor location. Third, to address the challenge of providing semantic context and meaning to indoor location detections, the framework uses this novel boosting methodology and proposes an approach to detect the floor-specific location and spatial information interpretation of a user's indoor position. Here, spatial information interpretation refers to detecting whether the specific user is located inside or outside the confines an indoor spatial region. These two functionalities of our framework were tested on a dataset that consisted of the user interaction data of multiple users during different activities in 3 buildings consisting of 5 floors and 254 spatial regions. The results presented and discussed show that these functionalities of our framework outperform all prior works in this field in terms of performance characteristics and operational features. By using such an exhaustive dataset consisting of 18 users with diverse user interaction characteristics, this work also aims to address the limitation in prior works in this field centered around the lack of data from diverse users in the training set of AAL-based machine learning systems.

As per the best knowledge of the authors, no prior work has been done in this field thus far that uses a similar approach. Future work would involve conducting real-time experiments as per IRB-approved protocols with more than 18 users to implement and deploy this framework in IoT-based simulated smart living environments for developing real-time personalized AAL experiences for the elderly in a way that enhances the trust and acceptance of such AAL systems.

**Author Contributions:** conceptualization, N.T.; methodology, N.T.; software, N.T.; validation, N.T.; formal analysis, N.T.; investigation, N.T.; resources, N.T.; data curation, N.T.; visualization, N.T.; data analysis and results, N.T.; writing—original draft preparation, N.T.; writing—review and editing, N.T.; supervision, C.Y.H.; project administration, C.Y.H. All authors have read and agreed to the published version of the manuscript.

**Funding:** This research was partially supported by the University of Cincinnati Graduate School Dean's Dissertation Completion Fellowship.

**Institutional Review Board Statement:** Not applicable.

**Informed Consent Statement:** Not applicable.

**Data Availability Statement:** Publicly available datasets were analyzed in this study. These data can be found at https://archive.ics.uci.edu/ml/datasets/ujiindoorloc, accessed on 13 December 2020.

**Acknowledgments:** The authors would like to thank Isabella Hall, Department of Electrical Engineering and Computer Science at the University of Cincinnati for her assistance in formatting and presentation of the different equations that constitute the proposed framework.

**Conflicts of Interest:** The authors declare no conflict of interest.

## References

1. Zhavoronkov, A.; Bischof, E.; Lee, K.-F. Artificial intelligence in longevity medicine. *Nat. Aging* **2021**, *1*, 5–7, https://doi.org/10.1038/s43587-020-00020-4.
2. Decade of Healthy Ageing (2021–2030). Available online: https://www.who.int/initiatives/decade-of-healthy-ageing (accessed on 28 June 2021).
3. WHO. Ageing and Health. Available online: https://www.who.Int/news-room/fact-sheets/detail/ageing-and-health (accessed on 20 December 2020).
4. Thakur, N.; Han, C. An Ambient Intelligence-Based Human Behavior Monitoring Framework for Ubiquitous Environments. *Information* **2021**, *12*, 81, https://doi.org/10.3390/info12020081.
5. Peine A, Marshall BL, Martin, W., Neven, L. (Eds.) *Socio-Gerontechnology: Interdisciplinary Critical Studies of Ageing and Technology*; Routledge: London, UK, 2021. Available online: https://play.google.com/store/books/details?id=QfkWEAAAQBAJ (accessed on 18 February 2021).
6. Chai, H.; Fu, R.; Coyte, P.C. Unpaid Caregiving and Labor Force Participation among Chinese Middle-Aged Adults. *Int. J. Environ. Res. Public Health* **2021**, *18*, 641, https://doi.org/10.3390/ijerph18020641.
7. Kadambi, S.; Loh, K.P.; Dunne, R.; Magnuson, A.; Maggiore, R.; Zittel, J.; Flannery, M.; Inglis, J.; Gilmore, N.; Mohamed, M.; et al. Older adults with cancer and their caregivers—current landscape and future directions for clinical care. *Nat. Rev. Clin. Oncol.* **2020**, *17*, 742–755, https://doi.org/10.1038/s41571-020-0421-z.
8. Flaherty, E.; Bartels, S.J. Addressing the Community-Based Geriatric Healthcare Workforce Shortage by Leveraging the Potential of Interprofessional Teams. *J. Am. Geriatr. Soc.* **2019**, *67*, S400–S408, https://doi.org/10.1111/jgs.15924.
9. Feng, Z.; Glinskaya, E.; Chen, H.; Gong, S.; Qiu, Y.; Xu, J.; Yip, W. Long-term care system for older adults in China: Policy landscape, challenges, and future prospects. *Lancet* **2020**, *396*, 1362–1372, https://doi.org/10.1016/s0140-6736(20)32136-x.
10. Javed, A.R.; Fahad, L.G.; Farhan, A.A.; Abbas, S.; Srivastava, G.; Parizi, R.M.; Khan, M.S. Automated cognitive health assessment in smart homes using machine learning. *Sustain. Cities Soc.* **2020**, *65*, 102572, https://doi.org/10.1016/j.scs.2020.102572.
11. United Nations. 68% of the World Population Projected to Live in Urban Areas by 2050, Says UN. 2018. Available online: https://www.un.org/development/desa/en/news/population/2018-revision-of-world-urbanization-prospects.html (accessed on 6 February 2021).
12. Furman, S., Haney, J. Is my home smart or just connected? In: *Artificial Intelligence in HCI*; Springer International Publishing: Cham, Switzerland, 2020; pp. 273–287.
13. Patro, S.P.; Padhy, N.; Chiranjevi, D. Ambient assisted living predictive model for cardiovascular disease prediction using supervised learning. *Evol. Intell.* **2020**, *14*, 941–969, https://doi.org/10.1007/s12065-020-00484-8.
14. Smirek, L.; Zimmermann, G.; Beigl, M. Just a Smart Home or Your Smart Home—A Framework for Personalized User Interfaces Based on Eclipse Smart Home and Universal Remote Console. *Procedia Comput. Sci.* **2016**, *98*, 107–116, https://doi.org/10.1016/j.procs.2016.09.018.
15. Marcus A, Aykin N, Chavan AL, Prabhu GV, Kurosu, M. SIG on one size fits all? Cultural diversity in user interface design. In CHI '99 Extended Abstracts On Human Factors In Computing Systems—CHI '99, Pittsburg, PY, USA, 15–20 May 1999; ACM Press: New York, NY, USA, 1999.
16. Iancu, I.; Iancu, B. Designing mobile technology for elderly. A theoretical overview. *Technol. Forecast. Soc. Chang.* **2020**, *155*, 119977, https://doi.org/10.1016/j.techfore.2020.119977.
17. Stonebraker, M., Çetintemel, U. "One size fits all": An idea whose time has come and gone. In *Making Databases Work: The Pragmatic Wisdom of Michael Stonebraker*; Association for Computing Machinery, New York, NY, USA, 2018. p. 441–462.

18. Yahya, M.A.; Dahanayake, A. A Needs-Based Augmented Reality System. *Appl. Sci.* **2021**, *11*, 7978, https://doi.org/10.3390/app11177978.
19. Imam, M., Savioz, P., O'Suilleabhain, C. IoT bridge components—Specialized smart monitoring solutions to address us-er-specific needs. In *Bridge Maintenance, Safety, Management, Life-Cycle Sustainability and Innovations*; CRC Press: Boca Raton, FL, USA, 2021; pp. 217–217.
20. Phillips, B.; Zhao, H. Predictors of Assistive Technology Abandonment. *Assist. Technol.* **1993**, *5*, 36–45, https://doi.org/10.1080/10400435.1993.10132205.
21. Dawe, M. Desperately seeking simplicity: How young adults with cognitive disabilities and their families adopt assistive technologies. In Proceedings of the SIGCHI Conference On Human Factors In Computing Systems—CHI '06, Montreal, QC, Canada, 22–27 April 2006; ACM Press: New York, NY, USA, 2006.
22. Gajos, K.Z.; Weld, D.S.; Wobbrock, J.O. Automatically generating personalized user interfaces with Supple. *Artif. Intell.* **2010**, *174*, 910–950, https://doi.org/10.1016/j.artint.2010.05.005.
23. Chaki, D.; Bouguettaya, A.; Mistry, S. A Conflict Detection Framework for IoT Services in Multi-resident Smart Homes. In Proceedings of the International Conference on Web Services, Beijing, China, 19–23 October 2020; pp. 224–231, https://doi.org/10.1109/icws49710.2020.00036.
24. Sikder AK, Babun L, Celik ZB, Acar A, Aksu, H., McDaniel, P.; et al. Kratos: Multi-user multi-device-aware access control system for the smart home. In Proceedings of the 13th ACM Conference on Security and Privacy in Wireless and Mobile Networks, Linz, Austria, 2020; ACM: New York, NY, USA, 2020.
25. Wilkowska, W., Ziefle, M. User diversity as a challenge for the integration of medical technology into future smart home environments. In *User-Driven Healthcare*; IGI Global: Hershey, PA, USA, 2012. p. 553–582.
26. Li, Q.; Gravina, R.; Li, Y.; Alsamhi, S.H.; Sun, F.; Fortino, G. Multi-user activity recognition: Challenges and opportunities. *Inf. Fusion* **2020**, *63*, 121–135, https://doi.org/10.1016/j.inffus.2020.06.004.
27. Thakur, N., Han CY. A review of assistive technologies for Activities of Daily Living of elderly. *arXiv* **2021**, Page: 26 arXiv:2106.12183
28. Kaptelinin, V.; Nardi, B. Activity Theory in HCI: Fundamentals and Reflections. *Synth. Lect. Hum.-Cent. Inform.* **2012**, *5*, 1–105, https://doi.org/10.2200/s00413ed1v01y201203hci013.
29. Allen, D.; Karanasios, S.; Slavova, M. Working with activity theory: Context, technology, and information behavior. *J. Am. Soc. Inf. Sci. Technol.* **2011**, *62*, 776–788, https://doi.org/10.1002/asi.21441.
30. Miller, H.J. Activity-Based Analysis. In *Handbook of Regional Science*; Springer: Berlin/Heidelber, Germany, 2021; pp. 187–207.
31. Thakur, N.; Han, C. Multimodal Approaches for Indoor Localization for Ambient Assisted Living in Smart Homes. *Information* **2021**, *12*, 114, https://doi.org/10.3390/info12030114.
32. Langlois, C.; Tiku, S.; Pasricha, S. Indoor Localization with Smartphones: Harnessing the Sensor Suite in Your Pocket. *IEEE Consum. Electron. Mag.* **2017**, *6*, 70–80, https://doi.org/10.1109/MCE.2017.2714719.
33. Varma, P.S.; Anand, V. Random Forest Learning Based Indoor Localization as an IoT Service for Smart Buildings. *Wirel. Pers. Commun.* **2020**, *117*, 3209–3227, https://doi.org/10.1007/s11277-020-07977-w.
34. Gao J, Li X, Ding Y, Su Q, Liu, Z. WiFi-based indoor positioning by random forest and adjusted cosine similarity. In Proceedings of the 2020 Chinese Control And Decision Conference (CCDC), Hefei, China, 21–23 May 2020; pp. 1426–1231.
35. Ullah Khan, I., Ali, T., Farid, Z., Scavino, E., Amiruddin Abd Rahman, M., Hamdi, M.; Qiao, G. An improved hybrid indoor posi-tioning system based on surface tessellation artificial neural network. *Meas. Control.* **2020**, *53*, 1968–1977.
36. Labinghisa, B.A.; Lee, D.M. Neural network-based indoor localization system with enhanced virtual access points. *J. Supercomput.* **2020**, *77*, 638–651, https://doi.org/10.1007/s11227-020-03272-4.
37. Qin, F.; Zuo, T.; Wang, X. CCpos: WiFi Fingerprint Indoor Positioning System Based on CDAE-CNN. *Sensors* **2021**, *21*, 1114, https://doi.org/10.3390/s21041114.
38. Musa, A.; Nugraha, G.D.; Han, H.; Choi, D.; Seo, S.; Kim, J. A decision tree-based NLOS detection method for the UWB indoor location tracking accuracy improvement: Decision-Tree NLOS Detection for the UWB Indoor Location Tracking. *Int. J. Commun. Syst.* **2019**, *32*, e3997.
39. Yim, J. Introducing a decision tree-based indoor positioning technique. *Expert Syst. Appl.* **2006**, *34*, 1296–1302, https://doi.org/10.1016/j.eswa.2006.12.028.
40. Sjoberg M, Koskela M, Viitaniemi, V., Laaksonen, J. Indoor location recognition using fusion of SVM-based visual classifiers. In Proceedings of the 2010 IEEE International Workshop on Machine Learning for Signal Processing, Kitilla, Finland, 29 August–1 September 2010.
41. Zhang S, Guo J, Wang W, Hu, J. Indoor 2.5D Positioning of WiFi Based on SVM. In Proceedings of the 2018 Ubiquitous Positioning, Indoor Navigation and Location-Based Services (UPINLBS), Wuhan, China, 22–23 March 2018.
42. Zhang L, Zhao C, Wang Y, Dai, L. Fingerprint-based indoor localization using weighted K-nearest neighbor and weighted signal intensity. In Proceedings of the 2nd International Conference on Artificial Intelligence and Advanced Manufacture, Manchester, UK, 15–17 October; ACM: New York, NY, USA, 2020.
43. Ge, X., Qu, Z. Optimization WIFI indoor positioning KNN algorithm location-based fingerprint. In 2016 7th IEEE International Conference on Software Engineering and Service Science (ICSESS), Beijing, China, 26–28 August 2016; pp. 135–137.
44. Hu, J.; Liu, D.; Yan, Z.; Liu, H. Experimental Analysis on Weight K-Nearest Neighbor Indoor Fingerprint Positioning. *IEEE Internet Things, J.* **2018**, *6*, 891–897, https://doi.org/10.1109/jiot.2018.2864607.

45. Zhang, Q.; Wang, Y. A 3D mobile positioning method based on deep learning for hospital applications. *EURASIP J. Wirel. Commun. Netw.* **2020**, *2020*, 1–15, https://doi.org/10.1186/s13638-020-01784-4.
46. Poulose, A.; Han, D.S. Hybrid Deep Learning Model Based Indoor Positioning Using Wi-Fi RSSI Heat Maps for Autonomous Applications. *Electronics* **2020**, *10*, 2, https://doi.org/10.3390/electronics10010002.
47. Ben Jamâa, M.; Koubâa, A.; Baccour, N.; Kayani, Y.; Al-Shalfan, K.; Jmaiel, M. EasyLoc: Plug-and-Play RSS-Based Localization in Wireless Sensor Networks. In *Studies in Computational Intelligence*; Springer: Berlin/Heidelber, Germany, 2013; pp. 77–98, https://doi.org/10.1007/978-3-642-39301-3_5.
48. Barsocchi, P.; Lenzi, S.; Chessa, S.; Furfari, F. Automatic virtual calibration of range-based indoor localization systems. *Wirel. Commun. Mob. Comput.* **2011**, *12*, 1546–1557, https://doi.org/10.1002/wcm.1085.
49. Bolic, M.; Rostamian, M.; Djuric, P.M. Proximity Detection with RFID: A Step Toward the Internet of Things. *IEEE Pervasive Comput.* **2015**, *14*, 70–76, https://doi.org/10.1109/mprv.2015.39.
50. Angermann, M.; Robertson, P. FootSLAM: Pedestrian simultaneous localization and mapping without exteroceptive sensors—hitchhiking on human perception and cognition. *Proc. IEEE Inst. Electr. Electron. Eng.* **2012**, *100*, 1840–1848.
51. Evennou, F.; Marx, F. Advanced Integration of WiFi and Inertial Navigation Systems for Indoor Mobile Positioning. *EURASIP J. Adv. Signal. Process.* **2006**, *2006*, 086706, https://doi.org/10.1155/asp/2006/86706.
52. Wang, H.; Lenz, H.; Szabo, A.; Bamberger, J.; Hanebeck, U. WLAN-Based Pedestrian Tracking Using Particle Filters and Low-Cost MEMS Sensors. In Proceedings of the 2007 4th workshop on positioning, navigation and communication, Hannover, Germany, 23–24 October 2007; pp. 1–7, https://doi.org/10.1109/wpnc.2007.353604.
53. Klingbeil, L.; Wark, T. A wireless sensor network for real-time indoor localisation and motion monitoring. In Proceedings of the 2008 International Conference on Information Processing in Sensor Networks (IPSN 2008), St. Louis, MI, USA, 22–24 April 2008; pp. 39–50.
54. Murata M, Ahmetovic D, Sato D, Takagi H, Kitani KM, Asakawa, C. Smartphone-based indoor localization for blind navigation across building complexes. In Proceedings of the 2018 IEEE International Conference on Pervasive Computing and Communications (PerCom), Athens, Greece, 19–23 March 2018.
55. Yoo, J.; Johansson, K.H.; Kim, H.J. Indoor Localization Without a Prior Map by Trajectory Learning From Crowdsourced Measurements. *IEEE Trans. Instrum. Meas.* **2017**, *66*, 2825–2835, https://doi.org/10.1109/tim.2017.2729438.
56. Kothari, N.; Kannan, B.; Glasgwow, E.D.; Dias, M.B. Robust Indoor Localization on a Commercial Smart Phone. *Procedia Comput. Sci.* **2012**, *10*, 1114–1120, https://doi.org/10.1016/j.procs.2012.06.158.
57. Qian, J.; Ma, J.; Ying, R.; Liu, P.; Pei, L. An improved indoor localization method using smartphone inertial sensors. In Proceedings of the International Conference on Indoor Positioning and Indoor Navigation, Montbeliard, France, 28–31 October 2013; pp. 1–7.
58. Fusco, G.; Coughlan, J.M. Indoor Localization Using Computer Vision and Visual-Inertial Odometry. *Comput. Help People Spec. Needs* **2018**, *10897*, 86–93, https://doi.org/10.1007/978-3-319-94274-2_13.
59. Chang, R.Y.; Liu, S.-J.; Cheng, Y.-K. Device-Free Indoor Localization Using Wi-Fi Channel State Information for Internet of Things. In Proceedings of the 2018, IEEE Global Communication Conference, Abu Dhabi, United Arab Emirates, 9–13 December 2018; pp. 1–7, https://doi.org/10.1109/glocom.2018.8647261.
60. Pathapati Subbu, K., Gozick, B., Dantu, R. Indoor localization through dynamic time warping. In Proceedings of the 2011 IEEE International Conference on Systems, Man, and Cybernetics, Anchorage, AK, USA, 9–12 October 2011; pp. 1639–1644.
61. Zhou, B.; Yang, J.; Li, Q. Smartphone-Based Activity Recognition for Indoor Localization Using a Convolutional Neural Network. *Sensors* **2019**, *19*, 621, https://doi.org/10.3390/s19030621.
62. Chen, J.; Ou, G.; Peng, A.; Zheng, L.; Shi, J. An INS/WiFi Indoor Localization System Based on the Weighted Least Squares. *Sensors* **2018**, *18*, 1458, https://doi.org/10.3390/s18051458.
63. Xu H, Yang Z, Zhou Z, Shangguan, L., Yi, K., Liu, Y. Indoor localization via multi-modal sensing on smartphones. In Proceedings of the 2016 ACM International Joint Conference on Pervasive and Ubiquitous Computing, Heidelberg, Germany, 12–16 September 2016; ACM: New York, NY, USA, 2016.
64. Wang H, Sen S, Elgohary A, Farid M, Youssef M, Choudhury RR. No need to war-drive: Unsupervised indoor localization. In Proceedings of the 10th International Conference on Mobile Systems, Applications, and Services—MobiSys '12, Wood Bay Lake, UK, 25–29 June 2012; ACM Press: New York, NY, USA, 2012; pp. 197–210.
65. Röbesaat, J.; Zhang, P.; Abdelaal, M.; Theel, O. An improved BLE indoor localization with Kalman-based fusion: An experimental study. *Sensors* **2017**, *17*, 951.
66. Yang, Z.; Feng, X.; Zhang, Q. Adometer: Push the Limit of Pedestrian Indoor Localization through Cooperation. *IEEE Trans. Mob. Comput.* **2014**, *13*, 2473–2483, https://doi.org/10.1109/tmc.2014.2329855.
67. Wu, C.; Yang, Z.; Liu, Y. Smartphones Based Crowdsourcing for Indoor Localization. *IEEE Trans. Mob. Comput.* **2014**, *14*, 444–457, https://doi.org/10.1109/tmc.2014.2320254.
68. Gu, F.; Khoshelham, K.; Shang, J.; Yu, F.; Wei, Z. Robust and Accurate Smartphone-Based Step Counting for Indoor Localization. *IEEE Sensors J.* **2017**, *17*, 3453–3460, https://doi.org/10.1109/jsen.2017.2685999.
69. Niu, X.; Zhang, Z.; Wang, A.; Liu, J.; Liu, S. Online learning-based WIFI radio map updating considering high-dynamic environmental factors. *IEEE Access* **2019**, *7*, 110074–110085.
70. Gkoufas, Y., Katrinis, K. Copernicus: A robust AI-centric indoor positioning system. In Proceedings of the 2018 International Conference on Indoor Positioning and Indoor Navigation (IPIN), La Cite, France, 24–27 September 2018; pp. 206–212.

71. Gkoufas, Y.; Braghin, S. Anatomy and deployment of robust AI-centric indoor positioning system. In Proceedings of the 2019 IEEE International Conference on Pervasive Computing and Communications Workshops (PerCom Workshops), Pisa, Italy, 21–25 March 2019. pp. 340–342.
72. Zemouri, S.; Magoni, D.; Zemouri, A.; Gkoufas, Y.; Katrinis, K.; Murphy, J. An Edge Computing Approach to Explore Indoor Environmental Sensor Data for Occupancy Measurement in Office Spaces. In Proceedings of the 2018 International Smart Cities Conference, Kansas City, MI, USA, 16–19 September 2018; pp. 1–8, https://doi.org/10.1109/isc2.2018.8656753.
73. Zemouri S, Gkoufas Y, Murphy, J. A machine learning approach to indoor occupancy detection using non-intrusive en-vironmental sensor data. In Proceedings of the 3rd International Conference on Big Data and Internet of Things—BDIOT 2019, Melbourn, Australia, 22–24 August 2019; ACM Press: New York, NY, USA, 2019.
74. Alsehly, F., Arslan, T.; Sevak, Z. Indoor positioning with floor determination in multi story buildings. In 2011 International Conference on Indoor Positioning and Indoor Navigation, Guimaeres, Portugal, 21–23 September 2011; pp. 1–7.
75. Campos, R.S.; Lovisolo, L.; Campos, M. Wi-Fi multi-floor indoor positioning considering architectural aspects and controlled computational complexity. *Expert Syst. Appl.* **2014**, *41*, 6211–6223, https://doi.org/10.1016/j.eswa.2014.04.011.
76. Sun, L.; Zheng, Z.; He, T.; Li, F. Multifloor Wi-Fi Localization System with Floor Identification. *Int. J. Distrib. Sens. Netw.* **2015**, *11*, https://doi.org/10.1155/2015/131523.
77. Haque, F.; Dehghanian, V.; Fapojuwo, A.O.; Nielsen, J. A Sensor Fusion-Based Framework for Floor Localization. *IEEE Sens. J.* **2018**, *19*, 623–631, https://doi.org/10.1109/jsen.2018.2852494.
78. Garcia, G.; Nahapetian, A. Wearable computing for image-based indoor navigation of the visually impaired. In Proceedings of the Conference on Wireless Health, Bethesda, MY, USA, 14–16 October 2015; ACM: New York, NY, USA, 2015; p. 17, https://doi.org/10.1145/2811780.2811959.
79. Ruan, C.; Yu, M.; He, X.; Song, B. An Indoor Floor Positioning Method Based on Smartphone's Barometer. In Proceedings of the 2018 Ubiquitous Positioning, Navigation and Location Based Services, Wuhan, China, 22–23 March 2018; pp. 1–9, https://doi.org/10.1109/upinlbs.2018.8559769.
80. Delahoz, Y.; Labrador, M.A. A real-time smartphone-based floor detection system for the visually impaired. In Proceedings of the 2017 IEEE International Symposium on Medical Measurements and Applications (MeMeA), Rochester, MN, USA, 8–10 May 2017; pp. 27–32.
81. Saguna, S.; Zaslavsky, A.; Chakraborty, D. Complex activity recognition using context-driven activity theory and activity signatures. *ACM Trans. Comput. Interact.* **2013**, *20*, 1–34, https://doi.org/10.1145/2490832.
82. Thakur, N.; Han, C.Y. Towards a Knowledge Base for Activity Recognition of Diverse Users. In *Human Interaction, Emerging Technologies and Future Applications III*; Springer International Publishing: Cham, Switzerland, 2020; pp. 303–308, https://doi.org/10.1007/978-3-030-55307-4_46.
83. Biggs, N. The roots of combinatorics. *Hist. Math.* **1979**, *6*, 109–136, https://doi.org/10.1016/0315-0860(79)90074-0.
84. Thakur, N. Framework for a Context Aware Adaptive Intelligent Assistant for Activities Of Daily Living [Internet]. University of Cincinnati Thesis. 2019. Available online: https://search.proquest.com/openview/a8a1e3b158d0a43068c95d0d8e2ed31d/1?pq-origsite=gscholar&cbl=18750&diss=y (accessed on 1 September 2021).
85. Mohammed, M.; Khan, M.B.; Bashier, E.B.M. *Machine Learning*; CRC Press: Boca Raton, FL, 2016.
86. Gusain, K.; Gupta, A.; Popli, B. Transition-Aware Human Activity Recognition Using eXtreme Gradient Boosted Decision Trees. In *Advanced Computing and Communication Technologies*; Springer: Singapore, 2017; pp. 41–49, https://doi.org/10.1007/978-981-10-4603-2_5.
87. Scheurer S, Tedesco S, Brown KN, O'Flynn, B. Human activity recognition for emergency first responders via body-worn inertial sensors. In Proceedings of the 2017 IEEE 14th International Conference on Wearable and Implantable Body Sensor Networks (BSN), Eindoven, The Netherlands, 9–12 May 2017; pp. 5–8.
88. Subasi, A.; Dammas, D.H.; Alghamdi, R.D.; Makawi, R.A.; Albiety, E.A.; Brahimi, T.; Sarirete, A. Sensor Based Human Activity Recognition Using Adaboost Ensemble Classifier. *Procedia Comput. Sci.* **2018**, *140*, 104–111, https://doi.org/10.1016/j.procs.2018.10.298.
89. Keally M, Zhou G, Xing G, Wu, J.; Pyles, A. PBN: Towards practical activity recognition using smartphone-based body sensor networks. In Proceedings of the 9th ACM Conference on Embedded Networked Sensor Systems—SenSys '11, Washington, DC, USA, 1–4 November 2011; ACM Press: New York, NY, USA, 2011.
90. Hastie, T.; Tibshirani, R.; Friedman, J. Boosting and Additive Trees. In *The Elements of Statistical Learning*; Springer: New York, NY, USA, 2009; pp. 1–51.
91. Madeh Piryonesi, S.; El-Diraby, T.E. Using machine learning to examine impact of type of performance indicator on flexible pavement deterioration modeling. *J. Infrastruct. Syst.* **2021**, *27*, 04021005.
92. Wikipedia Contributors. Gradient Boosting. Wikipedia, The Free Encyclopedia. 2021. Available online: https://en.wikipedia.org/w/index.php?title=Gradient_boosting&oldid=1036319866 (accessed on 1 August 2021).
93. Schapire, R.E. Explaining AdaBoost. In *Empirical Inference*; Springer: Berlin/Heidelberg, Germany, 2013. pp. 37–52.
94. Wikipedia Contributors. AdaBoost. Available online: https://en.wikipedia.org/w/index.php?title=AdaBoost&oldid=1015653726 (accessed on 5 April 2021).
95. Anguita, D.; Ghelardoni, L.; Ghio, A.; Oneto, L.; Ridella, S. The 'K' in K-fold Cross Validation. In Proceedings of the 20th European Symposium on Artificial Neural Networks, Computational Intelligence and Machine Learning (ESANN), Bruges, Belgium, 2–4 October 2012; pp. 441–446.

96. Wikipedia Contributors. Cross-Validation (Statistics) [Internet]. Wikipedia, The Free Encyclopedia. 2021. Available online: https://en.wikipedia.org/w/index.php?title=Cross-validation_(statistics)&oldid=1033842588 (accessed on 2 August 2021).
97. Mierswa, I.; Wurst, M.; Klinkenberg, R.; Scholz, M.; Euler, T. YALE: Rapid prototyping for complex data mining tasks. In Proceedings of the 12th ACM SIGKDD International Conference on Knowledge Discovery and Data Mining—KDD '06, Philadelphia, PA, USA, 20–23 August 2006; ACM Press: New York, NY, USA, 2006.
98. Hofmann, M.; Klinkenberg, R. *Rapidminer: Data Mining Use Cases and Business Analytics Applications*; CRC Press: Boca Raton, FL, USA, 2016.
99. Chakraborty S, Han CY, Zhou X, Wee WG. A context driven human activity recognition framework. In Proceedings of the 2016 International Conference on Health Informatics and Medical Systems, Monte Carlo Resort, Las Vegas, NV, USA, 25–28 July 2016; pp 96–102.
100. Center for Drug Evaluation, Research. (IRBs) and Protection of Human Subjects [Internet]. Fda.gov. 2019. Available online: https://www.fda.gov/about-fda/center-drug-evaluation-and-research-cder/institutional-review-boards-irbs-and-protection-human-subjects-clinical-trials (accessed on 11 January 2021).
101. nbsp;Axelrod, B. Ohio Gov. Mike DeWine Asks Employers to Continue Working Remotely Amid COVID-19. Available online: https://www.wkyc.com/article/news/health/coronavirus/dewine-woking-remotely-covid-19/95-54447569-e757-4eac-bbc3-4bab4652b764 (accessed on 5 April 2021).
102. Torres-Sospedra, J.; Montoliu, R.; Martinez-Uso, A.; Avariento, J.P.; Arnau, T.J.; Benedito-Bordonau, M.; Huerta, J. UJIIndoorLoc: A new multi-building and multi-floor database for WLAN fingerprint-based indoor localization problems. In Proceedings of the 2014 International Conference on Positioning and Indoor Navigation, Busan, Korea, 27–30 October 2014; pp. 261–270, https://doi.org/10.1109/ipin.2014.7275492.
103. Wikipedia Contributors. Jaume I University. Wikipedia, The Free Encyclopedia. 2021. Available online: https://en.wikipedia.org/w/index.php?title=Jaume_I_University&oldid=102570971 (accessed on 13 December 2020).
104. Wikipedia Contributors. Confusion Matrix. Wikipedia, The Free Encyclopedia. 2021. Available online: https://en.wikipedia.org/w/index.php?title=Confusion_matrix&oldid=1031861694 (accessed on 25 July 2021).
105. COVID-19 United States Cases by County—Johns Hopkins Coronavirus resource center. Available online: https://coronavirus.jhu.edu/us-map (accessed on 13 December 2020).
106. IHME COVID-19 Forecasting Team. Modeling COVID-19 scenarios for the United States. *Nat. Med.* **2021**, *27*, 94–105.